\pdfoutput=1
\documentclass[10pt,twocolumn,letterpaper]{article}

\usepackage{iccv}
\usepackage{times}
\usepackage{epsfig}
\usepackage{graphicx}
\usepackage{amsmath}
\usepackage{amssymb}
\usepackage{multirow}
\usepackage{float}
\usepackage{algorithm}
\usepackage[noend]{algpseudocode}
\usepackage{booktabs}
\usepackage{cuted}
\usepackage{capt-of}
\usepackage[title]{appendix}

\usepackage[pagebackref=true,breaklinks=true,letterpaper=true,colorlinks,bookmarks=false]{hyperref}

\iccvfinalcopy 

\def\iccvPaperID{****} 

\begin{document}

\newcommand{\ignore}[1]{}   
\newcommand{\cmt}[1]{\begin{sloppypar}\large\textcolor{red}{#1}\end{sloppypar}}
\newcommand{\note}[1]{\cmt{Note: #1}}

\newcommand{\todo}[1]{ \textcolor{red}{[{\bf TODO}: #1]}}
\newcommand{\torevise}[1]{\textcolor{blue}{#1}}
\newcommand{\copied}[1]{ \textcolor{red}{[COPIED: #1]}}
\newcommand{\frank}[1]{\textcolor{blue}{[Frank: #1]}}
\newcommand{\lulu}[1]{\textcolor{blue}{[Lulu: #1]}}

\newcommand{\comment}[1]{\textcolor{blue}{#1}}
\newcommand{\revised}[1]{\textcolor{blue}{#1}}
\newcommand{\ascalar}[1]{#1}
\newcommand{\avector}[1]{{\mathbf{#1}}}
\newcommand{\amatrix}[1]{\mathbf{#1}}
\newcommand{\aset}[1]{\mathbf{#1}}
\newcommand{\B}[1]{\mathbf{#1}}

\newcommand{\LL}{\mathcal{L}}
\newcommand{\E}{\mathbb{E}}
\newcommand{\Exp}[2]{\mathbb{E}_{#1} \left[ {#2} \right] }

\newcommand{\att}{a}
\newcommand{\A}{\mathbf{\att}}
\newcommand{\sourceA}{\A}
\newcommand{\targetA}{\hat{\A}}
\newcommand{\RA}{\mathbf{v}}
\newcommand{\coeff}{\alpha}

\newcommand{\x}{\mathbf{x}}
\newcommand{\y}{\mathbf{y}}

\newcommand{\adv}{\text{Real}}
\newcommand{\con}{\text{Match}}
\newcommand{\itp}{\text{Interp}}
\newcommand{\cyc}{\text{Cycle}}
\newcommand{\sel}{\text{Self}}

\newcommand{\Dadv}{D_\adv}
\newcommand{\Dcon}{D_\con}
\newcommand{\Dint}{D_\itp}

\newcommand{\Ladv}{\LL_\adv}
\newcommand{\Lcon}{\LL_\con}
\newcommand{\Lint}{\LL_\itp}
\newcommand{\Lcyc}{\LL_\cyc}
\newcommand{\Lsel}{\LL_\sel}

\newcommand{\w}{\lambda}

\newcommand{\zerovec}{\mathbf{0}}

\newcommand{\ourGAN}{RelGAN}

\title{\ourGAN: Multi-Domain Image-to-Image Translation via Relative Attributes}

\author{Po-Wei Wu$^1$~~~~Yu-Jing Lin$^1$~~~~Che-Han Chang$^2$~~~~Edward Y. Chang$^{2,3}$~~~~Shih-Wei Liao$^1$\\
$^1$National Taiwan University~~~~$^2$HTC Research \& Healthcare~~~~$^3$Stanford University\\
{\tt\scriptsize maya6282@gmail.com~~r06922068@ntu.edu.tw~~chehan\_chang@htc.com~~echang@cs.stanford.edu~~liao@csie.ntu.edu.tw}
}

\maketitle

\begin{abstract}
Multi-domain image-to-image translation has gained increasing attention recently. Previous methods take an image and some target attributes as inputs and generate an output image with the desired attributes. However, such methods have two limitations.  First, these methods assume binary-valued attributes and thus cannot yield satisfactory results for fine-grained control. Second, these methods require specifying the entire set of target attributes, even if most of the attributes would not be changed. To address these limitations, we propose RelGAN, a new method for multi-domain image-to-image translation. The key idea is to use relative attributes, which describes the desired change on selected attributes. Our method is capable of modifying images by changing particular attributes of interest in a continuous manner while preserving the other attributes. Experimental results demonstrate both the quantitative and qualitative effectiveness of our method on the tasks of facial attribute transfer and interpolation.
\end{abstract}

\section{Introduction} \label{introduction}

Multi-domain image-to-image translation aims to translate an image from one domain into another.
A domain is characterized by a set of attributes,
where each attribute is a meaningful property of an image.
Recently, this image-to-image translation problem received considerable attention following the emergence of generative adversarial networks (GANs)~\cite{goodfellow2014generative} and its conditional variants~\cite{mirza2014conditional}.
While most existing methods~\cite{isola2017image, yi2017dualgan, zhu2017unpaired, liu2017unsupervised} focus on image-to-image translation between two domains,
several multi-domain methods are proposed recently~\cite{choi2017stargan, he2019attgan, zhao2018modular}, which are capable of changing multiple attributes simultaneously.
For example, in the application of facial attribute editing, one can change hair color and expression simultaneously.

\begin{figure}
\begin{center}
	\includegraphics[width=1.0\linewidth]{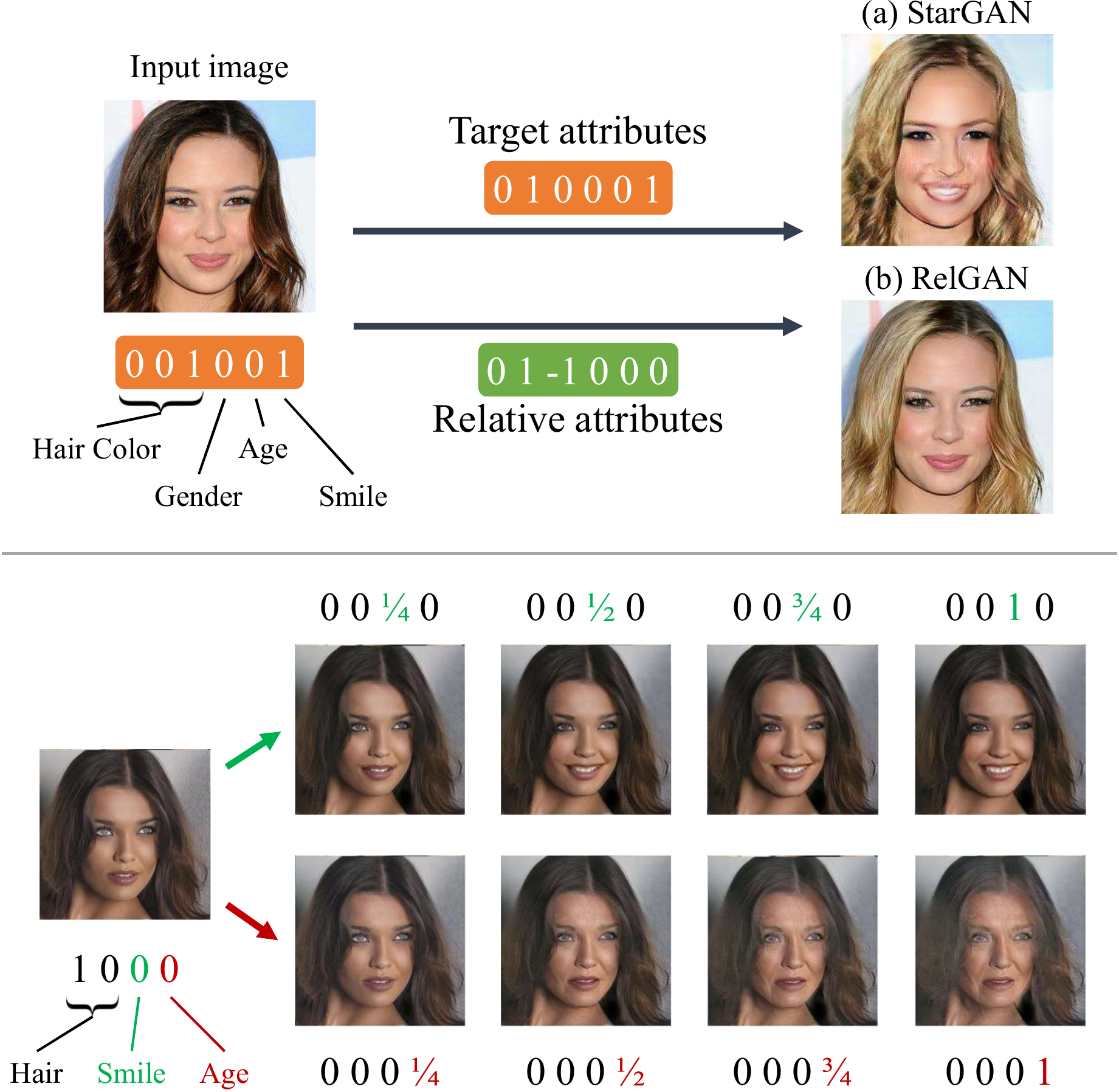}
	\caption{
	\textit{Top}: Comparing facial attribute transfer via relative and target attributes.
	(a) Existing target-attribute-based methods do not know whether each attribute is required to change or not, thus could over-emphasize some attributes.
	In this example, StarGAN changes the hair color but strengthens the degree of smile.
	(b) \ourGAN~only modifies the hair color and preserves the other attributes (including smile) because their relative attributes are zero.
	\textit{Bottom}:
	By adjusting the relative attributes in a continuous manner, \ourGAN~provides a realistic interpolation between before and after attribute transfer.
	}
	\label{fig:teaser}
	\vspace{-1em}
\end{center}
\end{figure}

\begin{figure*}
\begin{center}
	\includegraphics[width=1.0\linewidth]{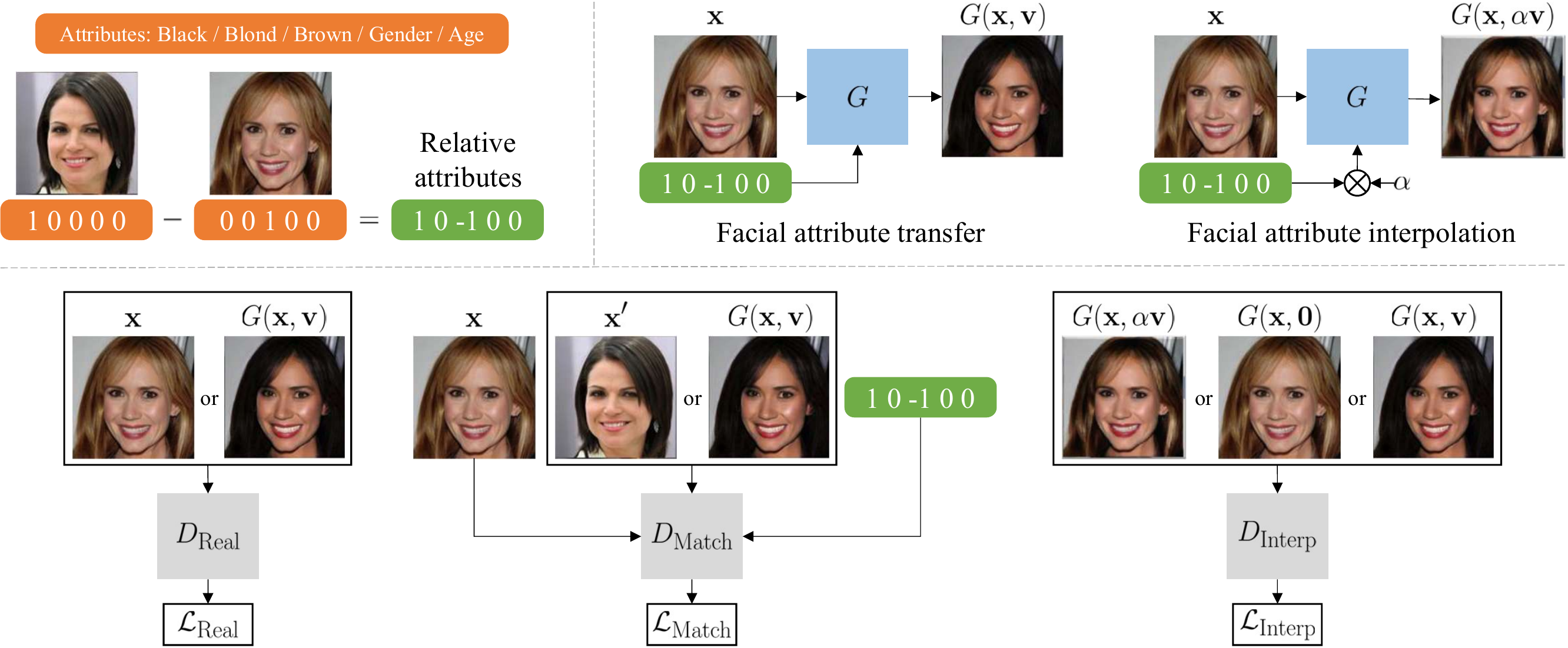}
	\caption{
	\textbf{\ourGAN}.
    Our model consists of a single generator $G$ and three discriminators $D = \{ \Dadv, \Dcon, \Dint \}$.
	$G$ conditions on an input image and relative attributes (\textit{top left}), and performs facial attribute transfer or interpolation (\textit{top right}).
	During training, 
	$G$ aims to fool the following three discriminators (\textit{bottom}):
	$\Dadv$ tries to distinguish between real images and generated images.
	$\Dcon$ aims to distinguish between real triplets and generated/wrong triplets. 
	$\Dint$ tries to predict the degree of interpolation.
	}
	\label{fig:overview}
\end{center}
\end{figure*}


Despite the impressive results of recent multi-domain methods~\cite{choi2017stargan, he2019attgan}, 
they have two limitations.
First, these methods assume binary attributes and therefore are not designed for attribute interpolation.
Although we can feed real-valued attributes into their generators, 
we found that their interpolation quality is unsatisfactory because their models trained on binary-valued attributes.
(Our model remedies
this shortcoming by training on real-valued relative attributes with 
additional discriminators.)
A smooth and realistic interpolation between before and after editing is important because it enables \textit{fine-grained control} over the strength of each attribute (e.g., the percentage of brown vs. blond hair color, or the degree of smile/happiness).

Second, these methods require a complete attribute representation to specify the target domain, 
even if only a subset of attributes is manipulated.
In other words, 
a user has to not only set the attributes of interest to the desired values but also identify the values of the unchanged attributes from the input image.
This poses a challenge for fine-grained control,
because a user does not know the underlying real value of each unchanged attribute.

To overcome these limitations, 
our key idea is that, unlike previous methods which take as input a pair $(\x, \targetA)$ of the original image $\x$ and target attributes $\targetA$, 
we take $(\x, \RA)$, where $\RA$ is the \textit{relative attributes} defined as the difference between the original attributes $\sourceA$ and the target ones $\targetA$, i.e., $\RA \triangleq \targetA - \sourceA$.
The values of the relative attributes directly encode how much each attribute is required to change.
In particular, non-zero values correspond to attributes of interest, while zero values correspond to unchanged attributes.
Figure~\ref{fig:teaser} illustrates our method with examples of facial attribute transfer and interpolation.
\ignore{
Notice that the upper interface requires the full set of
binary attributes being specified, while our
approach sets only those attributes requiring
changes with real value, which permit an
interpolation effect.
}

In this paper, we propose a relative-attribute-based method, dubbed \ourGAN, for multi-domain image-to-image translation.
\ourGAN~consists of a single generator $G$ and three discriminators $D = \{ \Dadv, \Dcon, \Dint \}$,
which are respectively responsible for guiding $G$ to learn to generate (1) realistic images, (2) accurate translations in terms of the relative attributes, and (3) realistic interpolations.
Figure~\ref{fig:overview} provides an overview of \ourGAN.



Our contributions can be summarized as follows:

\noindent 1. We propose \ourGAN, a relative-attribute-based method for multi-domain image-to-image translation.
\ourGAN~is based on the change of each attribute, and avoids the need to know the full attributes of an input image.

\noindent 2. To learn a generator conditioned on the relative attributes, 
we propose a matching-aware discriminator that determines whether an input-output pair matches the relative attributes.

\noindent 3. We propose an interpolation discriminator to improve the interpolation quality.

\noindent 4. We empirically demonstrate the effectiveness of \ourGAN~on facial attribute transfer and interpolation. Experimental results show that \ourGAN~achieves better results than state-of-the-art methods.





\section{Related Work}

We review works most related to ours and focus on conditional image generation and facial attribute transfer.
Generative adversarial networks (GANs)~\cite{goodfellow2014generative} are powerful unsupervised generative models that have gained significant attention in recent years.
Conditional generative adversarial networks (cGANs)~\cite{mirza2014conditional} extend GANs by conditioning both the generator and the discriminator on additional information.

Text-to-image synthesis and image-to-image translation can be treated as a cGAN that conditions on text and image, respectively.
For text-to-image synthesis, Reed et al.~\cite{reed2016generative} proposed a matching-aware discriminator to improve the quality of generated images.
Inspired by this work, we propose a matching-aware conditional discriminator.
StackGAN++~\cite{zhang2017stackgan++} uses a combination of an unconditional and a conditional loss as its adversarial loss.
For image-to-image translation, pix2pix~\cite{isola2017image} is a supervised approach based on cGANs.
To alleviate the problem of acquiring paired data for supervised learning, unpaired image-to-image translation methods~\cite{yi2017dualgan, zhu2017unpaired, liu2017unsupervised, huang2018munit} have recently received increasing attention.
CycleGAN~\cite{zhu2017unpaired}, the most representative method, learns two generative models and regularizes them by the cycle consistency loss.

Recent methods for facial attribute transfer~\cite{perarnau2016, lample2017fader, choi2017stargan, he2019attgan, zhao2018modular, pumarola2018ganimation} formulate the problem as unpaired multi-domain image-to-image translation.
IcGAN~\cite{perarnau2016} trains a cGAN and an encoder, and combines them into a single model that allows manipulating multiple attributes.
StarGAN~\cite{choi2017stargan} uses a single generator that takes as input an image and the target attributes to perform multi-domain image translation.
AttGAN~\cite{he2019attgan}, similar to StarGAN, performs facial attribute transfer based on the target attributes.
However, AttGAN uses an encoder-decoder architecture and treats the attribute information as a part of the latent representation, which is similar to IcGAN.
ModularGAN~\cite{zhao2018modular} proposes a modular architecture consisting of several reusable and composable modules.
GANimation~\cite{pumarola2018ganimation} trains its model on facial images with real-valued attribute labels and thus can achieve impressive results on facial expression interpolation.
\ignore{Although our model is trained on facial images with binary attributes, our interpolation is realistic and smoothly-varying.}

StarGAN~\cite{choi2017stargan} and AttGAN~\cite{he2019attgan} are two representative methods in multi-domain image-to-image translation.
\ourGAN~is fundamentally different from them in three aspects.
First, \ourGAN~employs a relative-attribute-based formulation rather than a target-attribute-based formulation.
Second, both StarGAN and AttGAN adopt auxiliary classifiers to guide the learning of image translation,
while \ourGAN's generator is guided by the proposed matching-aware discriminator, 
whose design follows the concept of conditional GANs~\cite{mirza2014conditional} and is tailored for relative attributes.
Third, we take a step towards continuous manipulation by incorporating an interpolation discriminator into our framework.
\section{Method}

In this paper, we consider that a domain is characterized by an $n$-dimensional attribute vector $\A = [\att^{(1)}, \att^{(2)}, \hdots, \att^{(n)}]^T$,
where each attribute $\att^{(i)}$ is a meaningful property of a facial image, such as age, gender, or hair color.
Our goal is to translate an input image $\x$ into an output image $\y$ such that $\y$ looks realistic and has the target attributes, where some user-specified attributes are different from the original ones while the other attributes remain the same.
To this end, we propose to learn a mapping function $(\x, \RA) \mapsto \y$, where $\RA$ is the relative attribute vector that represents the desired change of attributes.
Figure~\ref{fig:overview} gives an overview of \ourGAN.
In the following subsections, we first introduce relative attributes, then we describe the components of the \ourGAN~model.

\subsection{Relative Attributes} \label{sec:RA}

Consider an image $\x$, its attribute vector $\sourceA$ as the original domain, and the target attribute vector $\targetA$ as the target domain.
Both $\sourceA$ and $\targetA$ are $n$-dimensional vectors.
We define the relative attribute vector between $\sourceA$ and $\targetA$ as
\begin{align}
\RA \triangleq \targetA - \sourceA,
\end{align}
which naturally represents the desired change of attributes in modifying the input image $\x$ into the output image $\y$.

We argue that expressing the user's editing requirement by the relative attribute representation is straightforward and intuitive.
For example, if image attributes are binary-valued ($0$ or $1$), the corresponding relative attribute representation is three-valued ($-1$, $0$, $1$), where each value corresponds to a user's action to a binary attribute: turn on ($+1$), turn off ($-1$), or unchanged ($0$).
From this example, we can see that relative attributes encode the user requirement and have an intuitive meaning.

Next, facial attribute interpolation via relative attributes is rather straightforward:
to perform an interpolation between $\x$ and $G(\x, \RA)$,
we simply apply $G(\x, \coeff\RA)$, where $\coeff \in [0, 1]$ is an interpolation coefficient.

\subsection{Adversarial Loss}

We apply adversarial loss~\cite{goodfellow2014generative} to make the generated images indistinguishable from the real images.
The adversarial loss can be written as:
\begin{equation}
\begin{split}
\min_{G} \max _{\Dadv}
\Ladv & = \Exp{\x}{ \log \Dadv(\x) } \\
      & + \Exp{\x, \RA}{ \log(1-\Dadv(G(\x,\RA))) },
\end{split}
\label{eq:Lreal}
\end{equation}
where the generator $G$ tries to generate images that look realistic.
The discriminator $\Dadv$ is unconditional and aims to distinguish between the real images and the generated images.

\subsection{Conditional Adversarial Loss}

We require not only that the output image $G(\x, \RA)$ should look realistic, but also that the difference between $\x$ and $G(\x, \RA)$ should match the relative attributes $\RA$.
To achieve this requirement, we adopt the concept of conditional GANs~\cite{mirza2014conditional} and introduce a conditional discriminator $\Dcon$ that takes as inputs an image and the conditional variables, which is the pair $(\x, \RA)$.
The conditional adversarial loss can be written as:
\begin{equation}
\small
\begin{split}
\min_{G} \max _{\Dcon}
\Lcon & = \Exp{\x, \RA, \x'}{ \log \Dcon(\x, \RA, \x') } \\
      & + \Exp{\x, \RA}{ \log (1 - \Dcon(\x, \RA, G(\x, \RA))) }.
\end{split}
\label{eq:Lmatch}
\end{equation}
From this equation, we can see that $\Dcon$ takes a triplet as input.
In particular, $\Dcon$ aims to distinguish between two types of triplets: \textit{real} triplets $(\x, \RA, \x')$ and \textit{fake} triplets $(\x, \RA, G(\x, \RA))$.
A real triplet $(\x, \RA, \x')$ is comprised of two real images $(\x, \x')$ and the relative attribute vector $\RA=\A'-\A$, where $\A'$ and $\A$ are the attribute vector of $\x'$ and $\x$ respectively.
Here, we would like to emphasize that our training data is unpaired, i.e., $\x$ and $\x'$ are of \textit{different} identities with different attributes.

Inspired by the matching-aware discriminator~\cite{reed2016generative}, we propose to incorporate a third type of triplets: \textit{wrong} triplet, which consists of two real images with mismatched relative attributes.
By adding wrong triplets, $\Dcon$ tries to classify the real triplets as $+1$ (real and matched) while both the fake and the wrong triplets as $-1$ (fake or mismatched).
In particular, we create wrong triplets using the following simple procedure: given a real triplet expressed by $(\x, \A'-\A, \x')$, we replace one of these four variables by a new one to create a wrong triplet. By doing so, we obtain four different wrong triplets.
Algorithm~\ref{algo1} shows the pseudo-code of our conditional adversarial loss.

\begin{algorithm}[H] 
\caption{Conditional adversarial loss} \label{algo1}
\begin{algorithmic}[1] 
\Function{match\_loss}{$\x_1, \x_2, \x_3, \A_1, \A_2, \A_3$}
\State $\RA_{12}, \RA_{32}, \RA_{13} \gets \A_2 - \A_1, \A_2 - \A_3, \A_3 - \A_1$
\State $s_r \gets \Dcon(\x_1, \RA_{12}, \x_2)$ \{real triplet\}
\State $s_f \gets \Dcon(\x_1, \RA_{12}, G(\x_1, \RA_{12}))$ \{fake triplet\}
\State $s_{w_1} \gets \Dcon(\x_3, \RA_{12}, \x_2)$ \{wrong triplet\}
\State $s_{w_2} \gets \Dcon(\x_1, \RA_{32}, \x_2)$ \{wrong triplet\}
\State $s_{w_3} \gets \Dcon(\x_1, \RA_{13}, \x_2)$ \{wrong triplet\}
\State $s_{w_4} \gets \Dcon(\x_1, \RA_{12}, \x_3)$ \{wrong triplet\}
\State $\Lcon^D \gets (s_r-1)^2 + s_f^2 + \sum_{i=1}^4 s_{w_i}^2$
\State $\Lcon^G \gets (s_f-1)^2$
\State \Return $\Lcon^D, \Lcon^G$
\EndFunction
\end{algorithmic}
\end{algorithm}

\subsection{Reconstruction Loss}

By minimizing the unconditional and the conditional adversarial loss, $G$ is trained to generate an output image $G(\x,\RA)$ such that $G(\x,\RA)$ looks realistic and the difference between $\x$ and $G(\x,\RA)$ matches the relative attributes $\RA$.
However, there is no guarantee that $G$ only modifies those attribute-related contents while preserves all the other aspects from a low level (such as background appearance) to a high level (such as the identity of an facial image).
To alleviate this problem, we propose a cycle-reconstruction loss and a self-reconstruction loss to regularize our generator.

\noindent\textbf{Cycle-reconstruction loss.} We adopt the concept of cycle consistency~\cite{zhu2017unpaired} and require that $G(:,\RA)$ and $G(:,-\RA)$ should be the inverse of each other.
Our cycle-reconstruction loss is written as
\begin{equation}
\min_{G}
\Lcyc = \Exp{\x,\RA}{ \left\| G(G(\x,\RA), -\RA) - \x \right\|_1 }.
\end{equation}

\noindent\textbf{Self-reconstruction loss.} When the relative attribute vector is a zero vector $\zerovec$, which means that no attribute is changed, the output image $G(\x, \zerovec)$ should be as close as possible to $\x$.
To this end, we define the self-reconstruction loss as:
\begin{equation}
\min_{G}
\Lsel = \Exp{\x}{ \left\| G(\x, \zerovec) - x \right\|_1 },
\end{equation}
where $G$ degenerates into an auto-encoder and tries to reconstruct $\x$ itself.
We use L1 norm in both reconstruction losses.

\begin{figure*}
\begin{center}
	\includegraphics[width=1\linewidth]{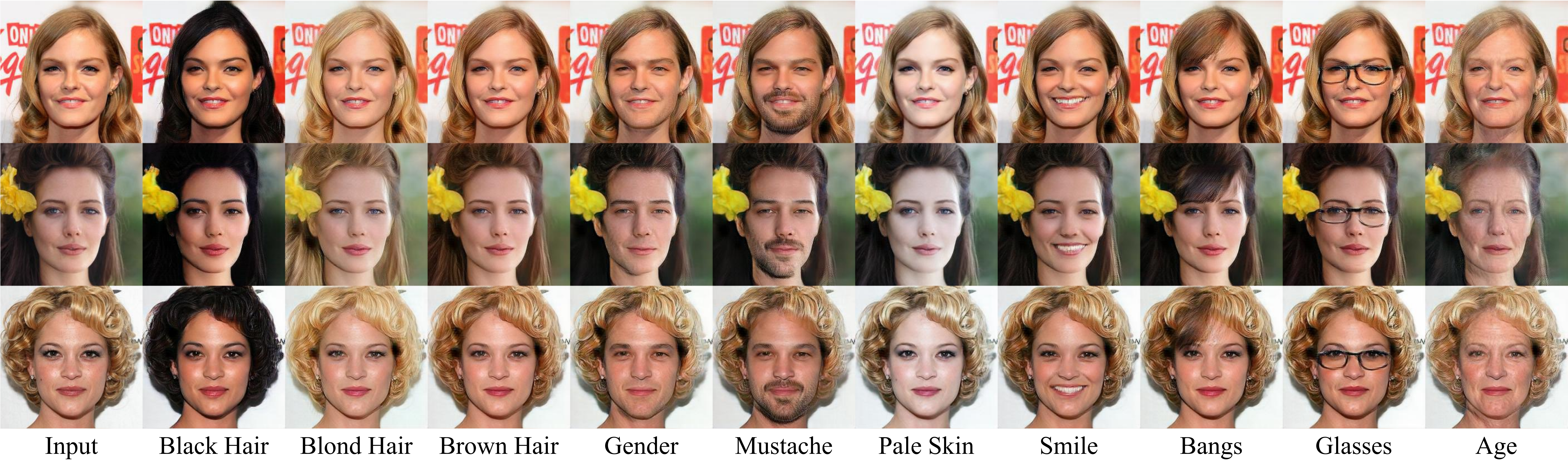}
	\caption{
	Facial attribute transfer results of \ourGAN~on the CelebA-HQ dataset.
	}
	\label{fig:big-teaser}
\end{center}
\end{figure*}

\subsection{Interpolation Loss}

Our generator interpolates between an image $\x$ and its translated one $G(\x, \RA)$ via $G(\x, \coeff\RA)$, where $\coeff$ is an interpolation coefficient.
To achieve a high-quality interpolation, we encourage the interpolated images $G(\x, \coeff\RA)$ to appear realistic.
Specifically, inspired by~\cite{berthelot2019understanding}, we propose a regularizer that aims to make $G(\x, \coeff\RA)$ indistinguishable from the non-interpolated output images, i.e., $G(\x, \zerovec)$ and $G(\x, \RA)$.
To this end, we introduce our third discriminator $\Dint$ to compete with our generator $G$.
The goal of $\Dint$ is to take an generated image as input and predict its \textit{degree of interpolation} $\hat{\coeff}$, which is defined as
$\hat{\coeff} = \min(\coeff, 1-\coeff)$,
where $\hat{\coeff}=0$ means no interpolation and $\hat{\coeff}=0.5$ means maximum interpolation.
By predicting $\hat{\coeff}$, we resolve the ambiguity between $\coeff$ and $1-\coeff$.

The interpolation discriminator $\Dint$ minimizes the following loss:
\begin{equation}
\begin{split}
\min_{\Dint}
\Lint^{D} = \mathbb{E}_{\x, \RA, \coeff} [ & \left\|\Dint(G(\x, \coeff\RA)) - \hat{\coeff}\right\|^2 \\
                                               + & \left\|\Dint(G(\x, \zerovec))\right\|^2 \\
                                               + & \left\|\Dint(G(\x, \RA))\right\|^2 ],
\end{split}
\end{equation}
where the first term aims at recovering $\hat{a}$ from $G(\x, \coeff\RA)$.
The second and the third term encourage $\Dint$ to output zero for the non-interpolated images.
The objective function of $G$ is modified by adding the following loss:
\begin{align}
\min_{G}
\Lint^{G} = \Exp{\x, \RA, \coeff}{ \left\| \Dint(G(\x, \coeff\RA)) \right\|^2 },
\end{align}
where $G$ tries to fool $\Dint$ to think that $G(\x, \coeff\RA)$ is non-interpolated.
In practice, we find empirically that the following modified loss stabilizes the adversarial training process:
\begin{equation}
\small
\begin{split}
\min_{\Dint}
\Lint^{D} = \mathbb{E}_{\x, \RA, \coeff} [ & \left\|\Dint(G(\x,\coeff\RA))-\hat{\coeff} \right\|^2 \\
+ & \left\|\Dint(G(\x,\mathbb{I}[\coeff>0.5]\RA))\right\|^2 ],
\end{split}
\end{equation}
where $\mathbb{I}[\cdot]$ is the indicator function that equals to $1$ if its argument is true and $0$ otherwise.
Algorithm~\ref{algo2} shows the pseudo-code of $\Lint^{D}$ and $\Lint^{G}$.

\begin{algorithm}[H]
\caption{Interpolation loss} \label{algo2}
\begin{algorithmic}[1]
\Function{interp\_loss}{$\x, \RA$}
\State $\alpha \sim \text{U}(0,1)$
\State $y_0 \gets \Dint(G(\x, \zerovec))$ \{non-interpolated image\}
\State $y_1 \gets \Dint(G(\x, \RA))$ \{non-interpolated image\}
\State $y_{\alpha} \gets \Dint(G(\x, \alpha \RA))$ \{interpolated image\}
\If{$\alpha \leq 0.5$}
    \State $\Lint^D \gets y_0^2 + (y_\alpha - \alpha)^2$
\Else
    \State $\Lint^D \gets y_1^2 + (y_\alpha - (1-\alpha))^2$
\EndIf
\State $\Lint^G \gets y_\alpha^2$
\State \Return $\Lint^D, \Lint^G$
\EndFunction
\end{algorithmic}
\end{algorithm}

\subsection{Full Loss}

To stabilize the training process,
we add orthogonal regularization~\cite{brock2018large} into our loss function.
Finally, the full loss function for $D = \{ \Dadv, \Dcon, \Dint \}$ and for $G$ are expressed, respectively, as

\begin{equation}
\begin{split}
\min_{D} \LL^{D} &= - \Ladv + \w_1 \Lcon^{D} + \w_2 \Lint^{D} \\
\end{split}
\label{eq:fullD}
\end{equation}
and
\begin{equation}
\begin{split}
\min_{G} \LL^{G} &=   \Ladv + \w_1 \Lcon^{G} + \w_2 \Lint^{G} \\
      &+ \w_3 \LL_{\text{Cycle}} + \w_4 \LL_{\text{Self}} + \w_5 \LL_{\text{Ortho}},
\end{split}
\label{eq:fullG}
\end{equation}
where $\w_1$, $\w_2$, $\w_3$, $\w_4$, and $\w_5$ are hyper-parameters that control the relative importance of each loss. 


\section{Experiments} \label{experiments}

In this section, we perform extensive experiments to demonstrate the effectiveness of \ourGAN.
We first describe the experimental settings (Section~\ref{sec41},~\ref{sec42}, and~\ref{sec43}).
Then we show the experimental results on the tasks of facial attribute transfer (Section~\ref{task:tsfer}),
facial image reconstruction (Section~\ref{task:recon}), 
and facial attribute interpolation (Section~\ref{task:interp}).
Lastly, we present the results of the user study (Section~\ref{task:user}).

\subsection{Dataset} \label{sec41}

\noindent\textbf{CelebA.}
The CelebFaces Attributes Dataset (CelebA)~\cite{liu2015faceattributes} contains $202{,}599$ face images of celebrities, annotated with $40$ binary attributes such as hair colors, gender, age.
We center crop these images to $178\times178$ and resize them to $256\times256$.

\noindent\textbf{CelebA-HQ.} Karras et al.~\cite{karras2017progressive} created a high-quality version of CelebA dataset, which consists of $30{,}000$ images generated by upsampling the CelebA images with an adversarially trained super-resolution model.

\noindent\textbf{FFHQ.}
The Flickr-Faces-HQ Dataset (FFHQ)~\cite{karras2019style} consists of $70{,}000$ high quality facial images at $1024\times 1024$ resolution.
This dataset has a larger variation than the CelebA-HQ dataset.

\begin{figure*}
\begin{center}
	\includegraphics[width=0.90\linewidth]{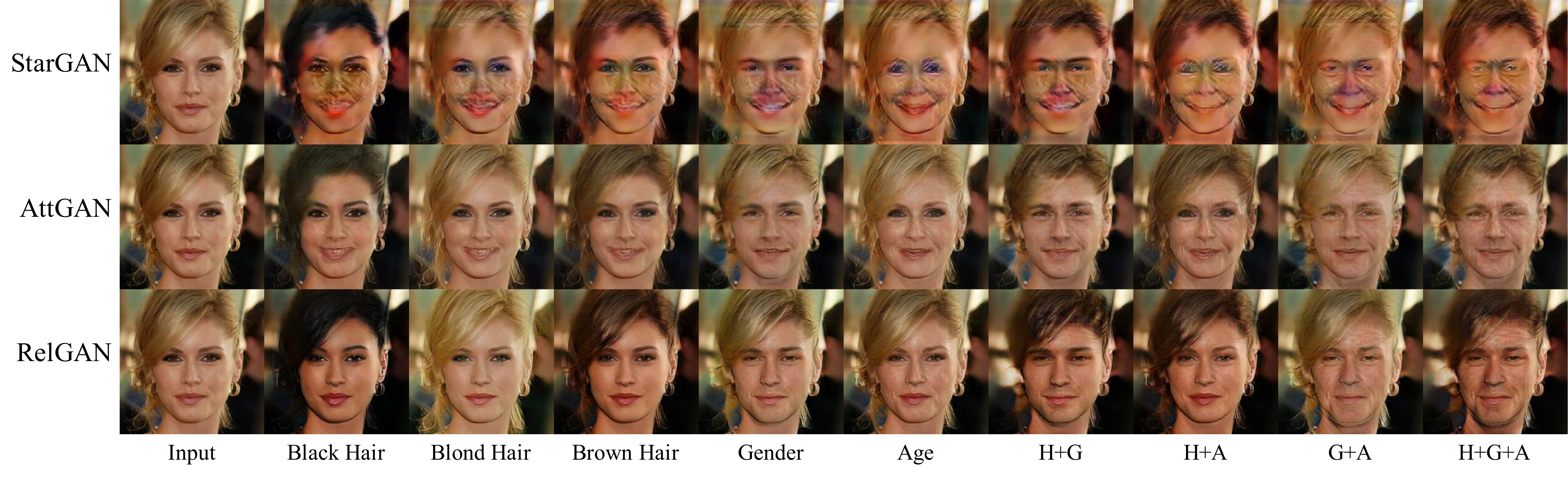}
	\caption{
	Facial attribute transfer results of StarGAN, AttGAN, and \ourGAN~on the CelebA-HQ dataset. Please zoom in for more details.
	}
	\label{fig:compare}
\end{center}
\end{figure*}

\subsection{Implementation Details} \label{sec42}
The images of the three datasets are center cropped and resized to $256\times 256$.
Our generator network, adapted from StarGAN~\cite{choi2017stargan}, is composed of two convolutional layers with a stride of $2$ for down-sampling, six residual blocks, and two transposed convolutional layers with a stride of $2$ for up-sampling.
We use switchable normalization~\cite{luo2018differentiable} in the generator.
Our discriminators $D = \{ \Dadv, \Dcon, \Dint \}$ have a shared feature sub-network comprised of six convolutional layers with a stride of $2$. Each discriminator has its output layers added onto the feature sub-network.
Please see the supplementary material for more details about the network architecture.

For $\Ladv$ (Equation~\ref{eq:Lreal}) and $\Lcon$ (Equation~\ref{eq:Lmatch}), 
we use LSGANs-GP~\cite{mao2018effectiveness} for stabilizing the training process.
For the hyper-parameters in Equation~\ref{eq:fullD} and~\ref{eq:fullG}, 
we use $\w_1=1$, $\w_2=\w_3=\w_4=10$, and $\w_5=10^{-6}$.
We use the Adam optimizer~\cite{kingma2014adam} with $\beta_1=0.5$ and $\beta_2=0.999$.
We train \ourGAN~from scratch with a learning rate of $5\times 10^{-5}$ and a batch size of $4$ on the CelebA-HQ dataset.
We train for $100$K iterations, which is about $13.3$ epochs.
Training \ourGAN~on a GTX $1080$ Ti GPU takes about $60$ hours.

\subsection{Baselines} \label{sec43}
We compare \ourGAN~with StarGAN~\cite{choi2017stargan} and AttGAN~\cite{he2019attgan},
which are two representative methods in multi-domain image-to-image translation.
For both methods, we use the code released by the authors and train their models on the CelebA-HQ dataset with their default hyper-parameters.

\begin{table}
    \footnotesize
	\centering
	\setlength{\tabcolsep}{4pt}
	\begin{tabular}{cccccc}
		\toprule
		Training set & $n$ & Test set & StarGAN & AttGAN & \ourGAN \\
		\midrule
		CelebA & 9 & CelebA & $10.15$ & $10.74$ & $\mathbf{4.68}$ \\
		CelebA-HQ & 9 & CelebA-HQ & $13.18$ & $11.73$ & $\mathbf{6.99}$ \\
		CelebA-HQ & 17 &CelebA-HQ &  $49.28$ & $13.45$ & $\mathbf{10.35}$ \\
		\midrule
		CelebA-HQ & 9 & FFHQ & $34.80$ & $25.53$ & $\mathbf{17.51}$ \\
		CelebA-HQ & 17 & FFHQ & $69.74$ & $27.25$ & $\mathbf{22.74}$ \\
		\bottomrule
	\end{tabular}
	\vspace{0.5em}
	\caption{
	\textbf{Visual quality comparison.}
	We use Fr\'echet Inception Distance (FID) for evaluating the visual quality (lower is better).
	$n$ is the number of attributes used in training.
	\ourGAN~achieves the lowest FID score among the three methods in all the five settings.
	}
	\label{table:fid}
\end{table}

\begin{table}
	\centering
	\small
	\setlength{\tabcolsep}{4pt}
	\begin{tabular}{ccccc}
		\toprule
		Images & Hair & Gender & Bangs & Eyeglasses\\
		\midrule
		CelebA-HQ & $92.52$ & $98.37$ & $95.83$ & $99.80$\\
		StarGAN & $\mathbf{95.48}$ & $90.21$ & $\mathbf{96.00}$ & $97.08 $\\
		AttGAN & $89.43$ & $94.44$ & $92.49$ & $98.26$\\
		\ourGAN & $91.08$ & $\mathbf{96.36}$ & $94.96$ & $\mathbf{99.20}$\\
		\toprule
		Images & Mustache & Smile & Pale Skin & Average\\
		\midrule
		CelebA-HQ & $97.90$ & $94.20$& $96.70$ & $96.47$\\
		StarGAN & $89.87$ & $90.56$ & $96.56$ & $93.68$\\
		AttGAN & $\mathbf{95.35}$ & $90.30$ & $\mathbf{98.23}$ & $94.07$\\
		\ourGAN & $94.57$ & $\mathbf{92.93}$ & $96.79$ & $\mathbf{95.13}$\\
		\bottomrule
	\end{tabular}
	\vspace{0.5em}
    \caption{
    The classification accuracy (percentage, higher is better) on the CelebA-HQ images and the generated images of StarGAN, AttGAN, and \ourGAN.
    For each attribute, the highest accuracy among the three methods is highlighted in bold.
    }
	\label{table:cls}
\end{table}

\begin{table*}
	\scriptsize
	\centering
	\begin{tabular}{cccc}
		\toprule
		$\Ladv$ & $\Lcon$ & $\Lcyc+\Lsel$ & Results \\
		\midrule
		$\surd$ & $\surd$ &         
		& \raisebox{-.5\height}{\includegraphics[width=0.6\linewidth]{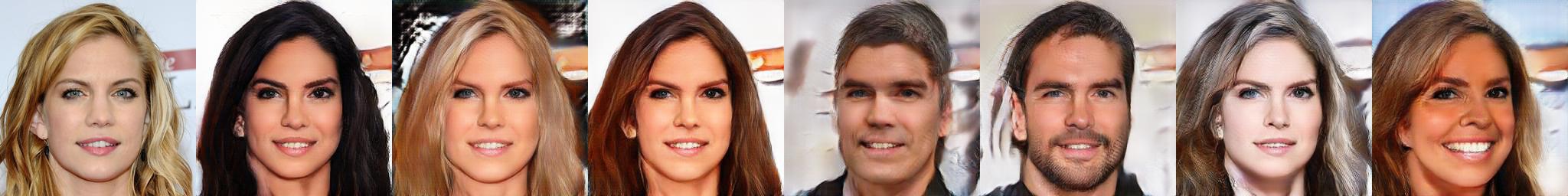}} \\
		\midrule
		$\surd$ &         & $\surd$	
		& \raisebox{-.5\height}{\includegraphics[width=0.6\linewidth]{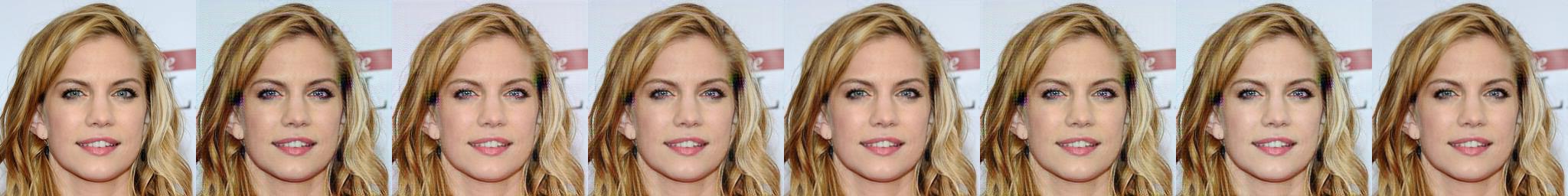}} \\
		\midrule
			    & $\surd$ & $\surd$	
	    & \raisebox{-.5\height}{\includegraphics[width=0.6\linewidth]{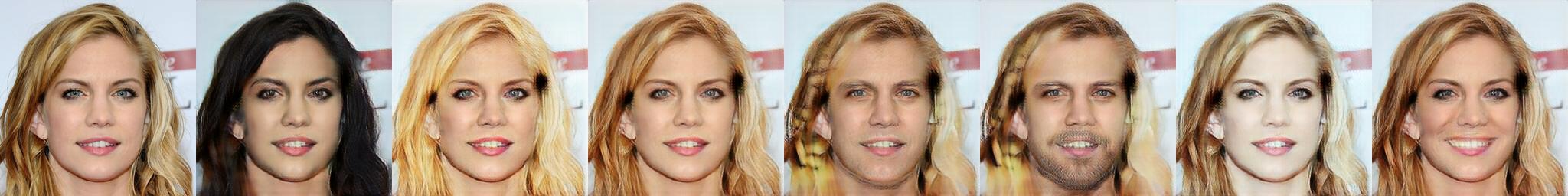}} \\
	    \midrule
		$\surd$ & $\surd$ & $\surd$ 
		& \raisebox{-.5\height}{\includegraphics[width=0.6\linewidth]{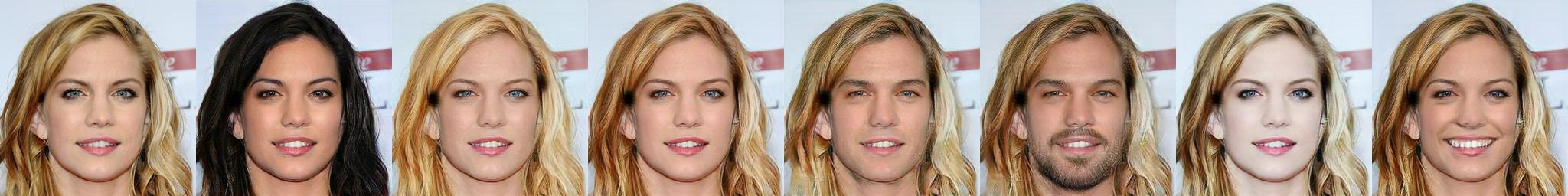}} \\
		\bottomrule
	\end{tabular}
	\vspace{0.5em}
	\caption{
	\textbf{Ablation study.} From left to right: input, black hair, blond hair, brown hair, gender, mustache, pale skin, and smile.
	}
	\label{table:ablation}
\end{table*}

\begin{figure*}
\begin{center}
	\includegraphics[width=0.90\linewidth]{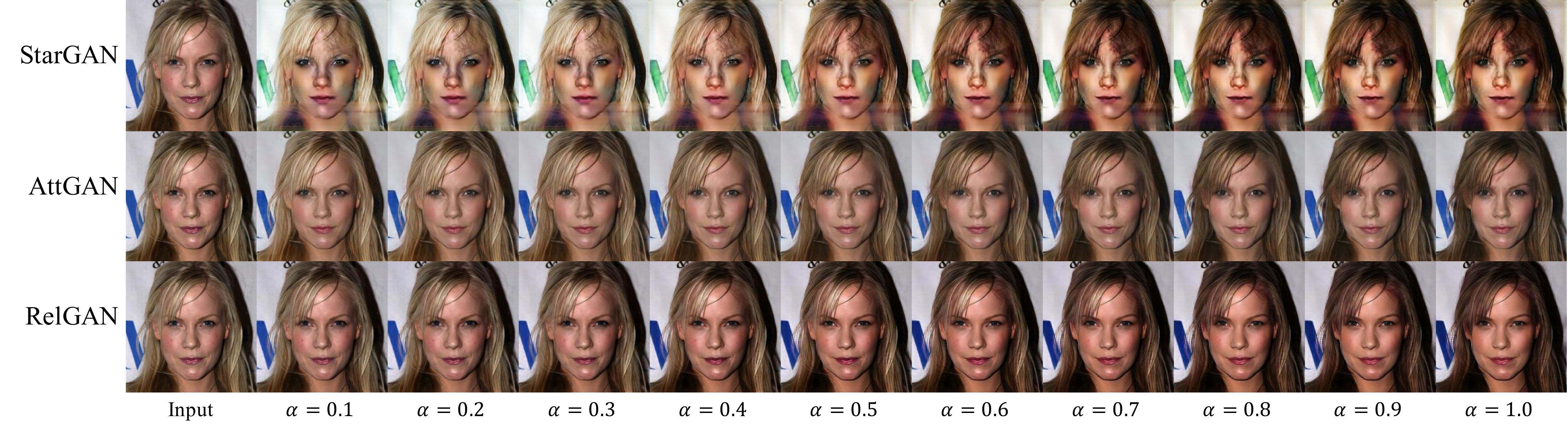}
	\caption{Facial attribute interpolation results of StarGAN, AttGAN, and \ourGAN~on the CelebA-HQ dataset.
	}
	\label{fig:interp_compare}
\end{center}
\end{figure*}

\begin{table}
	\small
	\centering
	\begin{tabular}{lccccc}
		\toprule
		Method & $\Lcyc$ & $\Lsel$ & L1 & L2 & SSIM \\
		\midrule
		StarGAN & $\surd$ & 		& $0.1136$ & $0.023981$ & $0.567$ \\
		AttGAN  & 		  & $\surd$	& $0.0640$ & $0.008724$ & $0.722$ \\
		\midrule
		\ourGAN	 	& $\surd$ & 		& $0.1116$ & $0.019721$ & $0.731$ \\
		\ourGAN 	& 		  & $\surd$ & $0.0179$ & $0.000649$ & $0.939$ \\
		\ourGAN 	& $\surd$ & $\surd$ & $\mathbf{0.0135}$ & $\mathbf{0.000463}$ & $\mathbf{0.947}$ \\
		\bottomrule
	\end{tabular}
	\vspace{0.5em}
	\caption{
	\textbf{Facial image reconstruction.}
	We measure the reconstruction error using L1 and L2 distance (lower is better), and SSIM (higher is better).
	}
	\label{table:recon_ssim}
\end{table}

\begin{table}
	\centering
	\begin{tabular}{lccc}
		\toprule
		Method & Hair & Age & Gender \\
		\midrule
		AttGAN & 				$0.0491$ & $0.0449$ & $0.0426$ \\
		StarGAN & 				$0.0379$ & $0.0384$ & $0.0375$ \\
		\ourGAN~w/o $\Lint$ &	$0.0363$ & $0.0308$ & $0.0375$ \\
		\ourGAN &			 	$\mathbf{0.0170}$ & $\mathbf{0.0278}$ & $\mathbf{0.0167}$ \\
		\bottomrule
	\end{tabular}
	\vspace{0.5em}
	\caption{\textbf{Facial attribute interpolation.}
	We measure the interpolation quality using the standard deviation of SSIM (Equation~\ref{eq:std}, lower is better).}
	\label{table:interp_ssim_std}
\end{table}

\begin{table*}
	\centering
	\setlength{\tabcolsep}{4pt}
	\begin{tabular}{c|cccccccc|cc}
		\toprule
		Method & Hair & Bangs & Eyeglasses  & Gender & Pale Skin & Smile  & Age & Mustache & Reconstruction & Interpolation\\
		\midrule
		StarGAN & $0.00$ & $0.74$ & $1.11$ & $1.11$ & $0.74$ & $2.21$ & $1.11$ & $0.74$ & $1.77$ & $6.05$\\
		AttGAN & $27.71$ & $34.32$ & $19.19$ & $28.78$ & $20.66$ & $\mathbf{52.76}$ & $42.44$ & $32.84$ & $7.82$ & $54.98$\\
		\ourGAN & $\mathbf{72.29}$ & $\mathbf{64.94}$ & $\mathbf{79.70}$ & $\mathbf{64.65}$ & $\mathbf{78.60}$ & $45.02$ & $\mathbf{56.46}$ & $\mathbf{66.42}$ & $\mathbf{97.12}$ & $\mathbf{66.42}$\\
		\bottomrule
	\end{tabular}
	\vspace{0.5em}
	\caption{
	The voting results of the user study (percentage, higher is better).
    }
	\label{table:userstudy}
\end{table*}

\subsection{Facial Attribute Transfer} \label{task:tsfer}

\noindent\textbf{Visual quality comparison.}
We use Fr\'echet Inception Distance (FID)~\cite{heusel2017gans} (lower is better) as the evaluation metric to measure the visual quality.
We experimented with three different training sets: CelebA with $9$ attributes, CelebA-HQ with $9$ attributes, and CelebA-HQ with $17$ attributes.
Table~\ref{table:fid} shows the FID comparison of StarGAN, AttGAN, and \ourGAN.
We can see that \ourGAN~consistently outperforms StarGAN and AttGAN for all the three training sets.
Additionally, 
we experimented with training on the CelebA-HQ dataset while testing on the FFHQ dataset to evaluate the generalization capability.
Still, \ourGAN~achieves better FID scores than the other methods.

\noindent\textbf{Classification accuracy.}
To quantitatively evaluate the quality of image translation,
we trained a facial attribute classifier on the CelebA-HQ dataset using a Resnet-18 architecture~\cite{he2016deep}.
We used a $90/10$ split for training and testing.
In Table~\ref{table:cls}, We report the classification accuracy on the test set images and the generated images produced by StarGAN, AttGAN, and \ourGAN.
The accuracy on the CelebA-HQ images serves as an upper-bound.
\ourGAN~achieves the highest average accuracy and rank 1st in $3$ out of the $7$ attributes.

\noindent\textbf{Qualitative results.}
Figure~\ref{fig:big-teaser} and~\ref{fig:compare} show the qualitative results on facial attribute transfer.
Figure~\ref{fig:big-teaser} shows representative examples to demonstrate that \ourGAN~is capable of generating high quality and realistic attribute transfer results.
Figure~\ref{fig:compare} shows a visual comparison of the three methods.
StarGAN's results contain notable artifacts.
AttGAN yields blurry and less detailed results compared to \ourGAN.
Conversely, \ourGAN~is capable of preserving unchanged attributes.
In the case of changing hair color, \ourGAN~preserves the smile attribute, while both StarGAN and AttGAN make the woman's mouth open due to their target-attribute-based formulation.
More qualitative results can be found in the supplementary material.

In Table~\ref{table:ablation}, 
we show an ablation study of our loss function.
We can see that:
(1) Training without $\Lcyc+\Lsel$ (1st row) cannot preserve identity.
(2) Training without $\Lcon$ (2nd row) only learns to reconstruct the input image.
(3) Training without $\Ladv$ (3rd row) gives reasonable results.
(4) Training with full loss (4th row) yields the best results.

\subsection{Facial Image Reconstruction} \label{task:recon}

One important advantage of \ourGAN~is preserving the unchanged attributes, which is a desirable property for facial attribute editing.
When all the attributes are unchanged, i.e., the target attribute vector is equal to the original one, the facial attribute transfer task reduces to a reconstruction task.
Here, we evaluate the performance of facial image reconstruction as a proxy metric to demonstrate that \ourGAN~better preserves the unchanged attributes.

To perform facial image reconstruction, 
we respectively apply StarGAN and AttGAN by taking the original attributes as the target attributes, 
and apply \ourGAN~by taking a zero vector as the relative attributes.
We measure L1, L2 norm, and SSIM similarity~\cite{wang2004image} between the input and the output images.
As shown in Table~\ref{table:recon_ssim},
StarGAN uses a cycle-reconstruction loss only while AttGAN uses a self-reconstruction loss only.
We evaluate three variants of \ourGAN~to uncover the contribution of $\Lcyc$ and $\Lsel$.
The results show that \ourGAN~without $\Lcyc$ already outperforms StarGAN and AttGAN in terms of all the three metrics.
\ourGAN~further improves the results.

\subsection{Facial Attribute Interpolation} \label{task:interp}

We next evaluate \ourGAN~on the task of facial attribute interpolation.
For both StarGAN and AttGAN, their interpolated images are generated by $G\left( \x, \coeff\sourceA+(1-\coeff)\targetA \right)$, where $\sourceA$ and $\targetA$ are the original and the target attribute vector, respectively.
Our interpolated images are generated by $G(\x, \coeff\RA)$.

\noindent\textbf{Qualitative results.}
As can be seen from Figure~\ref{fig:interp_compare}, StarGAN generates a non-smooth interpolation that the appearance change mainly happens between $\coeff=0.4$ and $\coeff=0.6$.
Both StarGAN and AttGAN have an abrupt change between the input and the result with $\coeff=0.1$.
In particular, the blond hair attribute is not well preserved by both methods.
\ourGAN~achieves the most smooth-varying interpolation.

\begin{figure}
\begin{center}
	\includegraphics[width=1\linewidth]{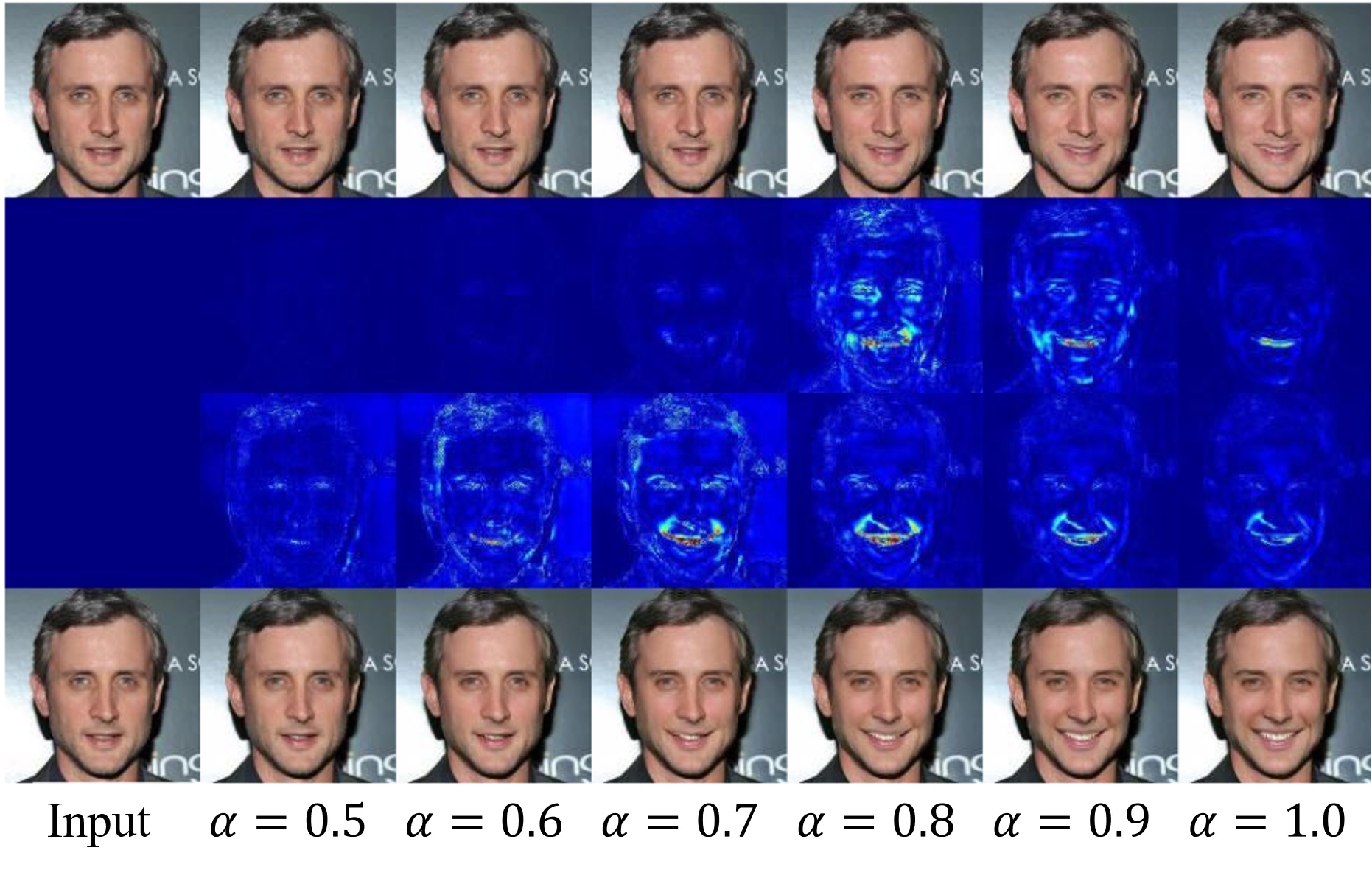}
	\caption{
	We use heat maps to visualize the difference between two adjacent images.
	\textbf{Top two rows}: interpolation without $\Lint$ gives an inferior interpolation due to an abrupt change between $\alpha=0.7$ and $\alpha=0.8$.
	\textbf{Bottom two rows}: interpolation with $\Lint$ gives better results since the appearance change is more evenly distributed across the image sequence.
	}
	\label{fig:interp_heat}
\end{center}
\end{figure}

\noindent\textbf{Quantitative evaluation.}
We use the following metric to evaluate the interpolation quality.
Given an input image $\x_0$, an output image $\x_m$, and a set of interpolated images $\left\{ \x_1, \cdots, \x_{m-1} \right\}$, a high quality and smoothly-varying interpolation implies that the appearance changes steadily from $\x_0$ to $\x_m$.
To this end, 
we compute the standard deviation of the SSIM scores between $\x_{i-1}$ and $\x_{i}$, i.e.,
\begin{align}
\sigma( \left\{\text{SSIM}(\x_{i-1}, \x_i)|i=1, \cdots, m\right\} ), \label{eq:std}
\end{align}
where $\sigma\left(\cdot\right)$ computes the standard deviation.
We use $m=10$ in this experiment.
A smaller standard deviation indicates a better interpolation quality.
As shown in Table~\ref{table:interp_ssim_std}, 
StarGAN is comparable to \ourGAN~without $\Lint$.
\ourGAN~with $\Lint$ effectively reduces the standard deviation, 
showing that our interpolation is not only realistic but also smoothly-varying.
Figure~\ref{fig:interp_heat} shows a visual comparison of \ourGAN~with and without $\Lint$.

\subsection{User Study} \label{task:user}

We conducted a user study to evaluate the image quality of \ourGAN.
We consider $10$ tasks, 
where eight are the facial attribute transfer tasks (Section~\ref{task:tsfer}), 
one is the facial image reconstruction (Section~\ref{task:recon}),
and one is the facial attribute interpolation (Section~\ref{task:interp}).
$302$ users were involved in this study. 
Each user is asked to answer $40$ questions,
each of which is generated from randomly sampling a Celeba-HQ image and a task, and then applying StarGAN, AttGAN, and \ourGAN~to obtain their result respectively.
For the attribute transfer tasks,
users are asked to pick the best result among the three methods.
For the other two tasks,
we allow users to vote for multiple results that look satisfactory.  
The results of the user study are summarized in Table~\ref{table:userstudy}.
\ourGAN~obtains the majority of votes in all the tasks except the smile task.

\section{Conclusion}

In this paper, we proposed a novel multi-domain image-to-image translation model based on relative attributes.
By taking relative attributes as input, our generator learns to modify an image in terms of the attributes of interest while preserves the other unchanged ones.
Our model achieves superior performance over the state-of-the-art methods in terms of both visual quality and interpolation.
Our future work includes using more advanced adversarial learning methods~\cite{jolicoeur2018relativistic, miyato2018cgans,chang2019kggan} and mask mechanisms~\cite{pumarola2018ganimation, zhao2018modular, zhang2018generative, sun2018mask} for further improvement.
{\small
\bibliographystyle{IEEEtran}
\bibliography{egbib}
}

\begin{appendices}
\onecolumn
\section{Network Architecture}

\begin{figure}[h]
    \centering
    \includegraphics[width=0.94\linewidth]{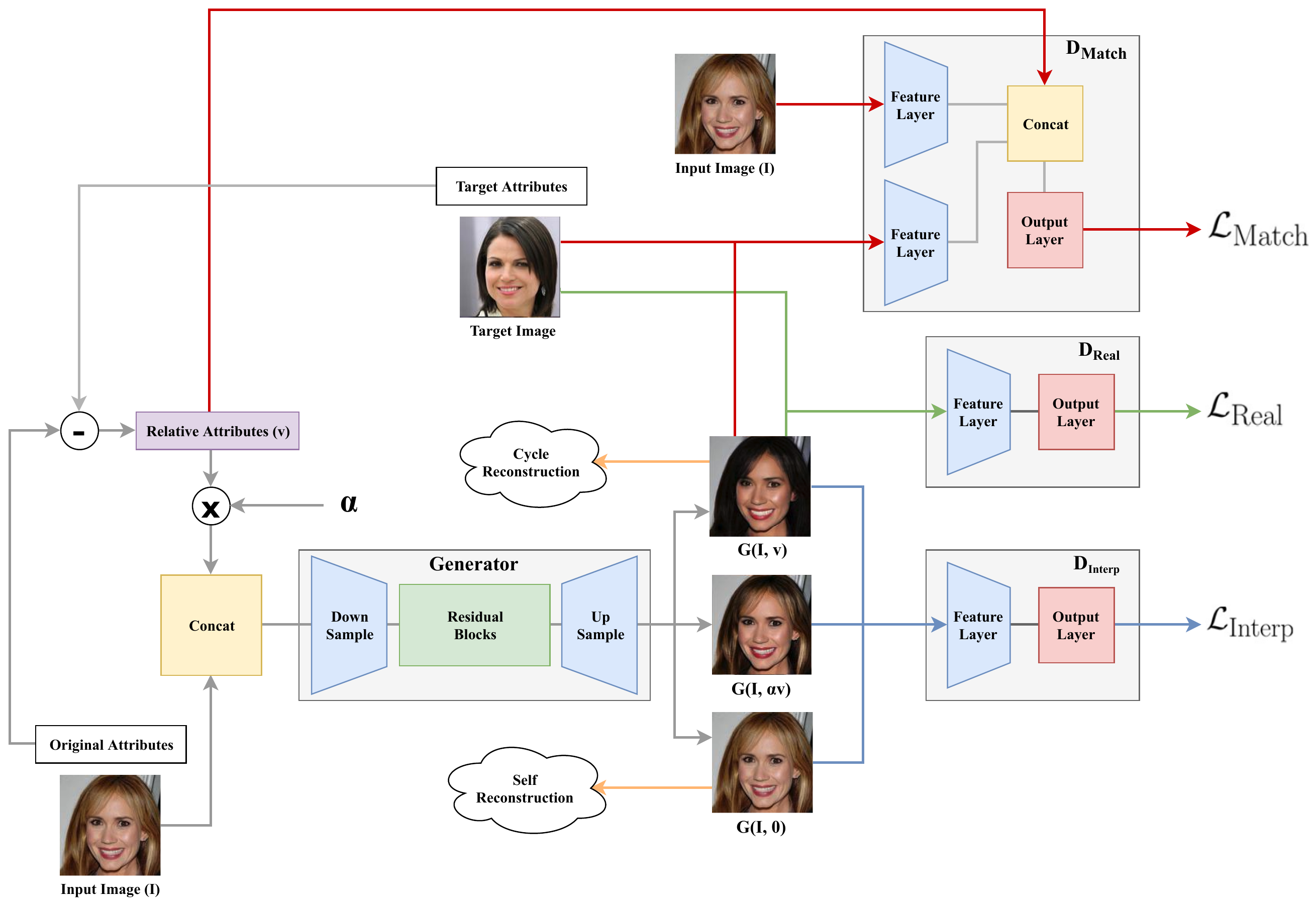}
    \caption{
    Detailed schematic diagram of \ourGAN.
    $\Dadv$, $\Dcon$, and $\Dint$ share the weights of the feature layers.
    }
    \label{fig:full-loss}
\end{figure}

Figure~\ref{fig:full-loss} shows the schematic diagram of \ourGAN.
Table~\ref{tab:generator} and~\ref{tab:discriminator} show the network architecture of \ourGAN.

\begin{table}[h]
\renewcommand\arraystretch{1.7}
    \centering
    \begin{tabular}{lcc}
        Component &  Input $\rightarrow$ Output Shape & Layer Information\\
        \hline
        \hline
        \multirow{3}{9em}{Down Sample} 
        & $(h,w,3+n)\rightarrow(h,w,64)$ & Conv-(F=64,K=7,S=1,P=3),SN,ReLU \\ 
        & $(h,w,64)\rightarrow(h,w,128)$ & Conv-(F=128,K=4,S=2,P=1),SN,ReLU \\ 
        & $(h,w,128)\rightarrow(h,w,256)$ & Conv-(F=256,K=4,S=2,P=1),SN,ReLU \\ 
        \hline
        \multirow{6}{9em}{Residual Blocks} 
        & $(h,w,256)\rightarrow(h,w,256)$ & Residual Block: Conv-(F=256,K=3,S=1,P=1),SN,ReLU \\
        & $(h,w,256)\rightarrow(h,w,256)$ & Residual Block: Conv-(F=256,K=3,S=1,P=1),SN,ReLU \\
        & $(h,w,256)\rightarrow(h,w,256)$ & Residual Block: Conv-(F=256,K=3,S=1,P=1),SN,ReLU \\
        & $(h,w,256)\rightarrow(h,w,256)$ & Residual Block: Conv-(F=256,K=3,S=1,P=1),SN,ReLU \\
        & $(h,w,256)\rightarrow(h,w,256)$ & Residual Block: Conv-(F=256,K=3,S=1,P=1),SN,ReLU \\
        & $(h,w,256)\rightarrow(h,w,256)$ & Residual Block: Conv-(F=256,K=3,S=1,P=1),SN,ReLU \\
        \hline
        \multirow{3}{9em}{Up Sample} 
        & $(h,w,256)\rightarrow(h,w,128)$ & Conv-(F=128,K=4,S=2,P=1),SN,ReLU \\ 
        & $(h,w,128)\rightarrow(h,w,64)$ & Conv-(F=64,K=4,S=2,P=1),SN,ReLU \\ 
        & $(h,w,64)\rightarrow(h,w,3)$ & Conv-(F=3,K=7,S=1,P=3),Tanh \\ 
    \end{tabular}
    \caption{
    \textbf{Generator network architecture.}
    We use switchable normalization, denoted as SN, in all layers except the last output layer.
    $n$ is the number of attributes.
    F is the number of filters.
    K is the filter size.
    S is the stride size.
    P is the padding size.
    The number of trainable parameters is about $8$M.
    }
    \label{tab:generator}
\end{table}

\begin{table}[h]
\renewcommand\arraystretch{1.7}
    \centering
    \begin{tabular}{lcc}
        Component &  Input $\rightarrow$ Output Shape & Layer Information\\
        \hline
        \hline
        \multirow{6}{9em}{Feature Layers} 
        & $(h,w,3)\rightarrow(h/2,w/2,64)$ & Conv-(F=64,K=4,S=2,P=1),Leaky ReLU \\ 
        & $(h/2,w/2,64)\rightarrow(h/4,w/4,128)$ & Conv-(F=128,K=4,S=2,P=1),Leaky ReLU \\
        & $(h/4,w/4,128)\rightarrow(h/8,w/8,256)$ & Conv-(F=256,K=4,S=2,P=1),Leaky ReLU \\
        & $(h/8,w/8,256)\rightarrow(h/16,w/16,512)$ & Conv-(F=512,K=4,S=2,P=1),Leaky ReLU \\
        & $(h/16,w/16,512)\rightarrow(h/16,w/16,1024)$ & Conv-(F=1024,K=4,S=2,P=1),Leaky ReLU \\
        & $(h/32,w/32,1024)\rightarrow(h/64,w/64,2048)$ & Conv-(F=2048,K=4,S=2,P=1),Leaky ReLU \\
        \hline 
        \multirow{1}{9em}{$\Dadv$: Output Layer} &  $(h/64,w/64,2048)\rightarrow(h/64,w/64,1)$ & Conv-(F=1,K=1)\\
        \hline
        \multirow{2}{9em}{$\Dint$: Output Layer} & $(h/64,w/64,2048)\rightarrow(h/64,w/64,64)$ & Conv-(F=64,K=1) \\
        & $(h/64,w/64,64)\rightarrow(h/64,w/64,1)$ & Mean(axis=3) \\
        \hline
        \multirow{2}{9em}{$\Dcon$: Output Layer} & $(h/64,w/64,4096+n)\rightarrow(h/64,w/64,2048)$ & Conv-(F=1,K=1),Leaky ReLU \\
        & $(h/64,w/64,2048)\rightarrow(h/64,w/64,1)$ & Conv-(F=1,K=1)\\
    \end{tabular}
    \caption{
    \textbf{Discriminator network architecture.}
    We use Leaky ReLU with a negative slope of $0.01$.
    $n$ is the number of attributes.
    F is the number of filters.
    K is the filter size.
    S is the stride size.
    P is the padding size.
    The number of trainable parameters is about $53$M.
    }
    \label{tab:discriminator}
\end{table}
\clearpage

\section{Additional Results}
\begin{figure}[h]
    \centering
    \includegraphics[width=0.7\linewidth]{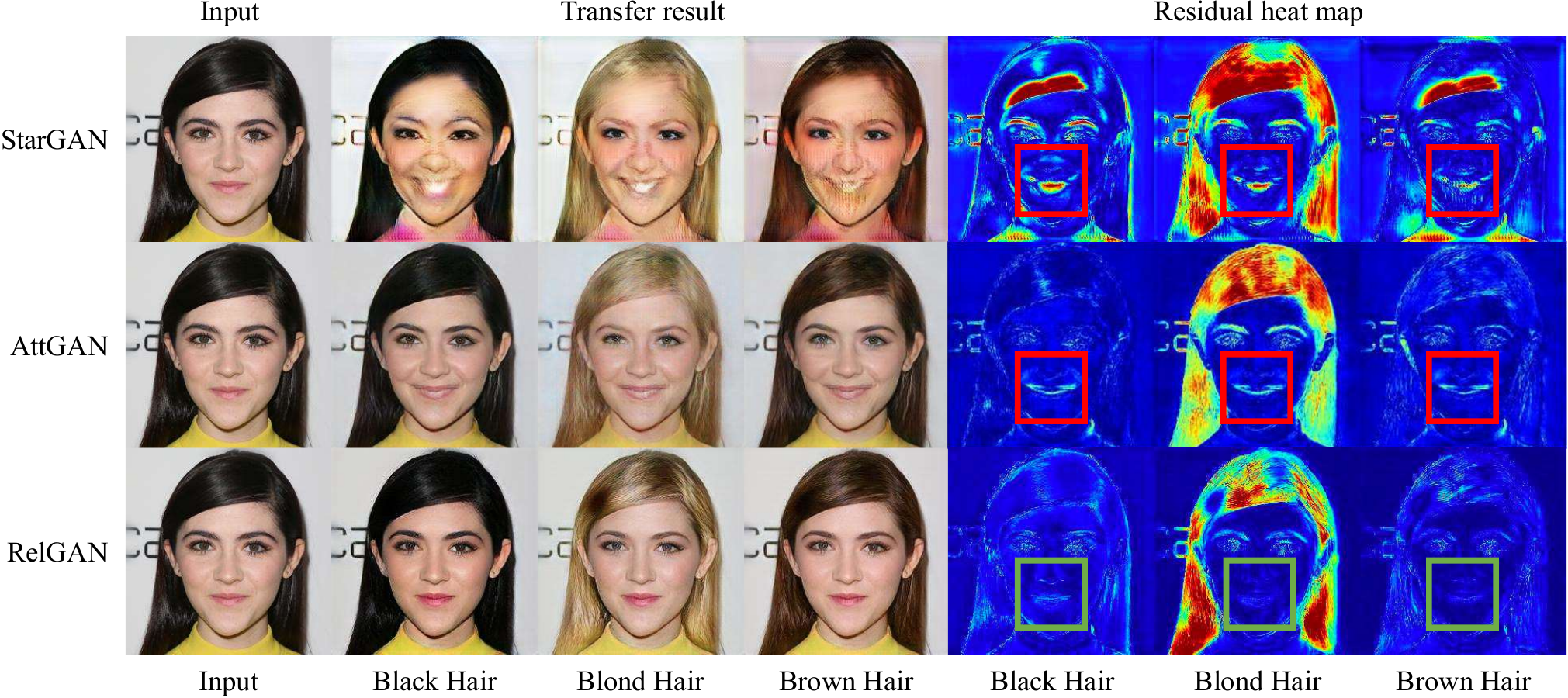}
    \caption{
    Hair color transfer results of StarGAN, AttGAN, and \ourGAN.
    The residual heat maps visualize the differences between the input and the output images.
    }
    \label{fig:residual1}
\end{figure}

\begin{figure}[h]
    \centering
    \includegraphics[width=0.7\linewidth]{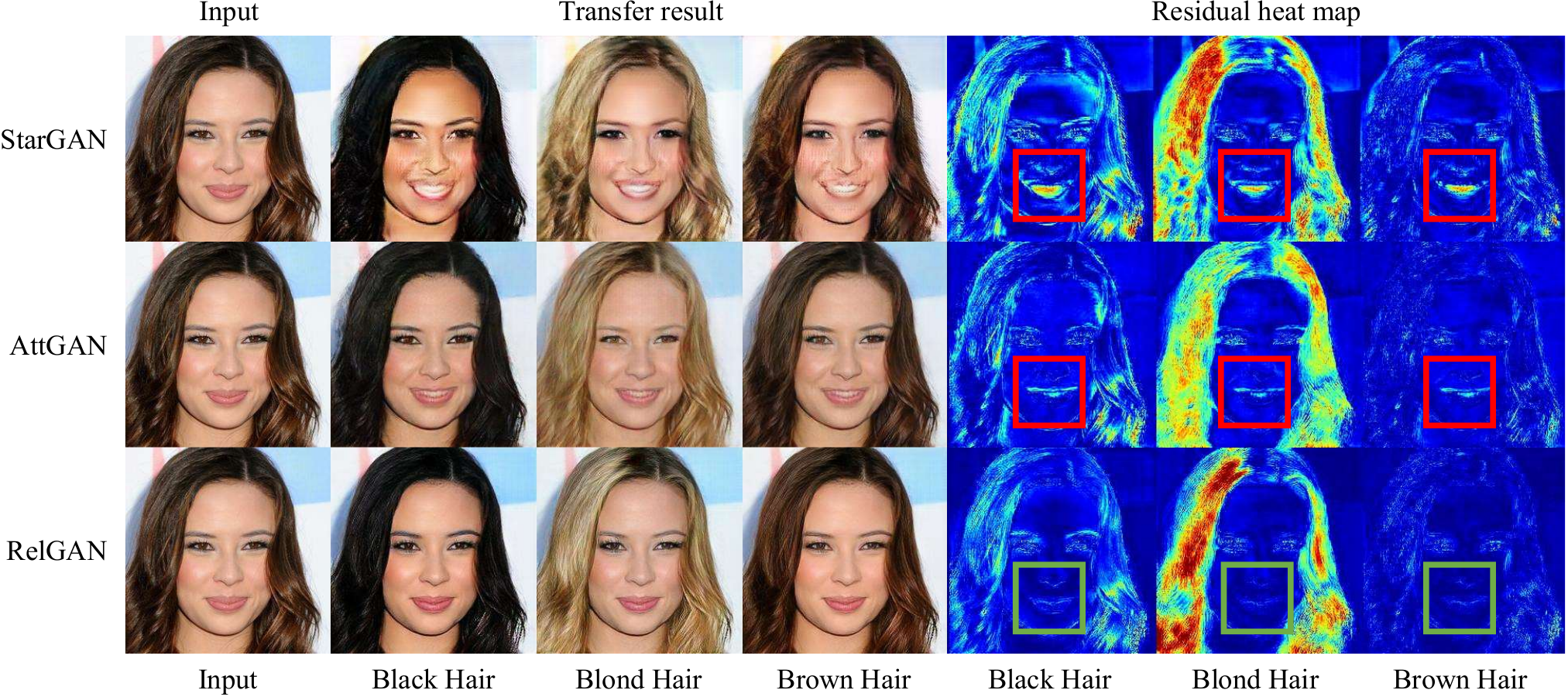}
    \caption{
    Hair color transfer results of StarGAN, AttGAN, and \ourGAN.
    The residual heat maps visualize the differences between the input and the output images.
    }
    \label{fig:residual2}
\end{figure}

\begin{figure}[h]
    \centering
    \includegraphics[width=1\linewidth]{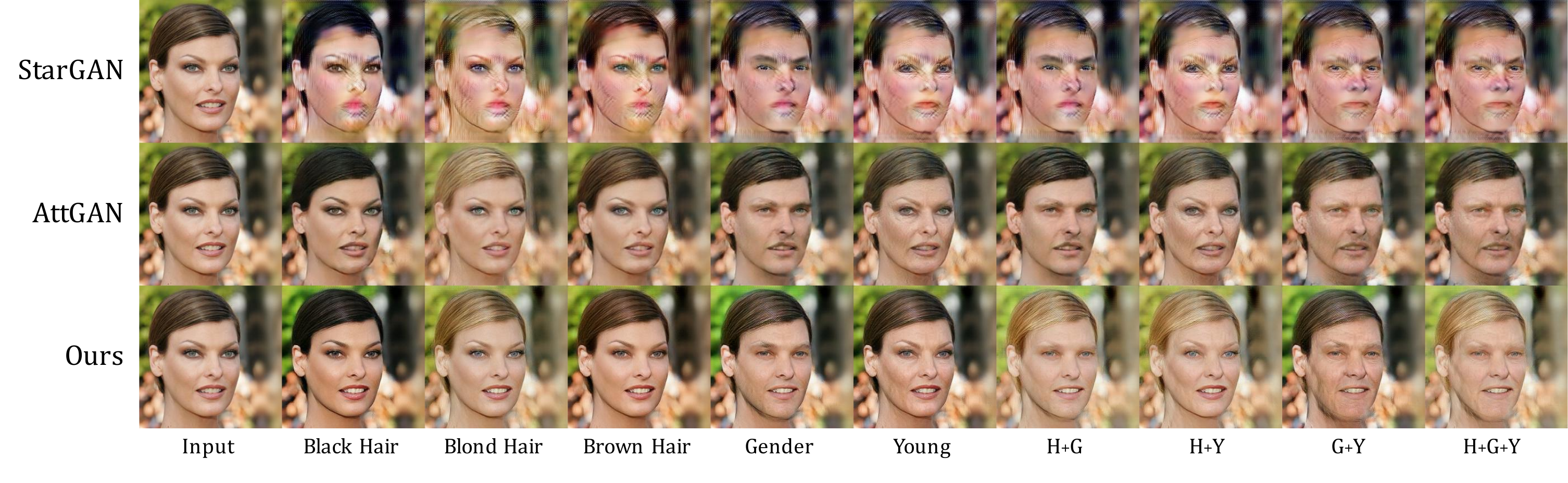}
    \vspace{-0.7cm}
    \caption{
    Facial attribute transfer results of StarGAN, AttGAN, and \ourGAN.
    Please zoom in for more details.
    In the case of changing hair color, \ourGAN~preserves the smile attribute, while both StarGAN and AttGAN make the woman look unhappy due to their target-attribute-based formulation.
    }
    \label{fig:model_compare}
\end{figure}

\begin{figure}[h]
    \centering
    \includegraphics[width=1\linewidth]{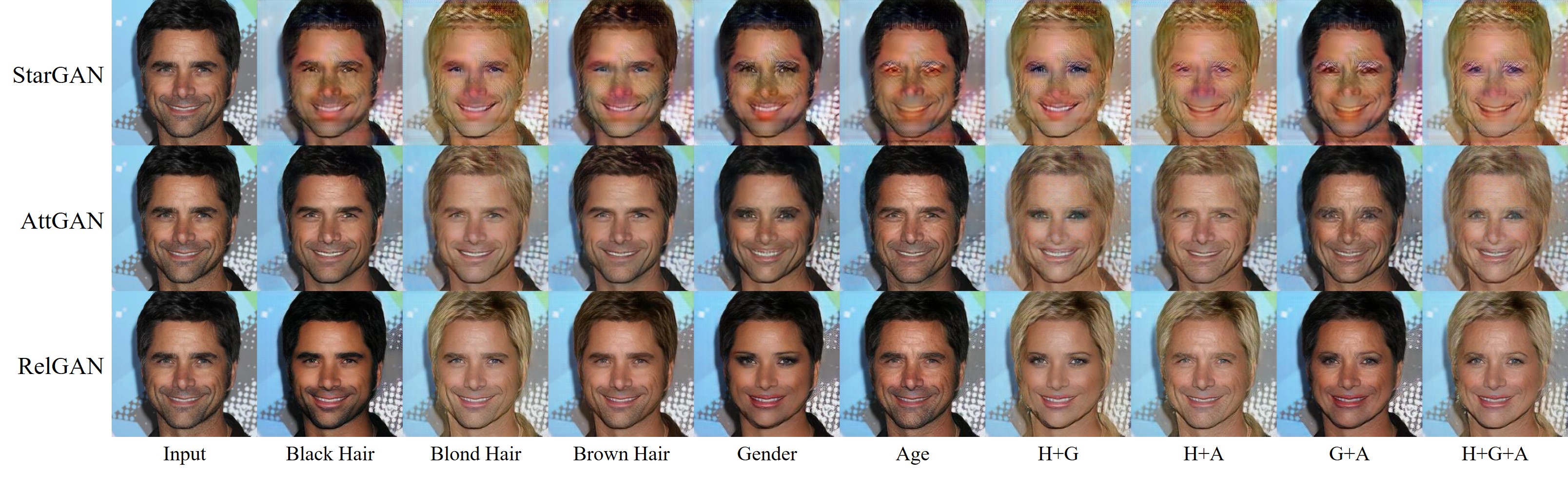}
    \includegraphics[width=1\linewidth]{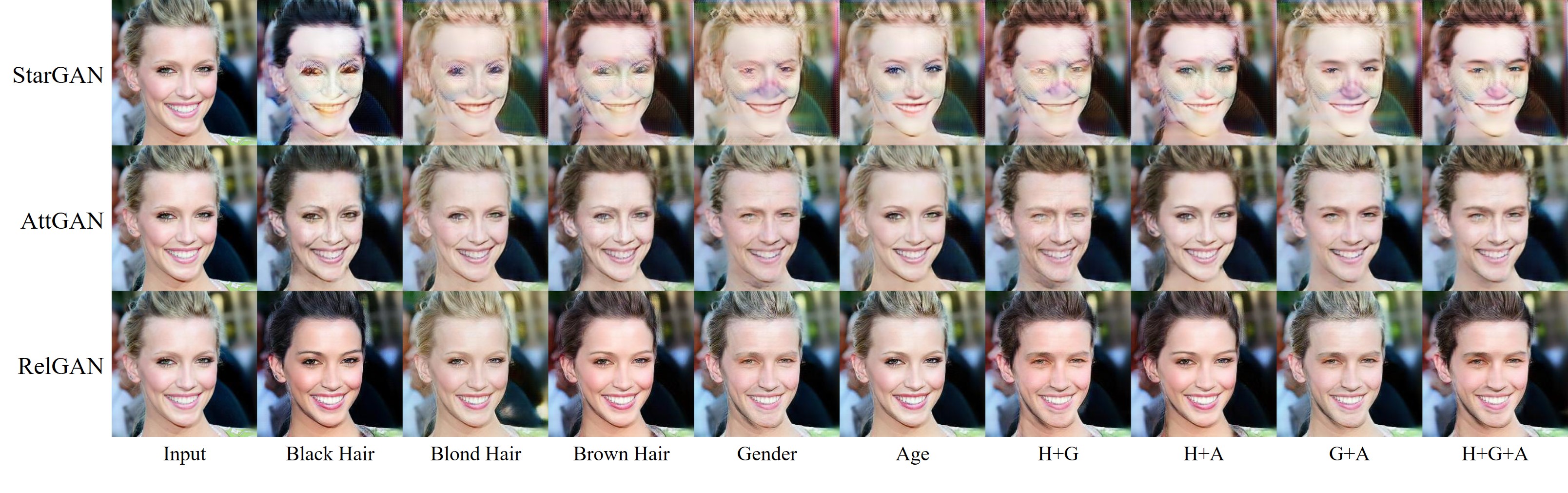}
    \includegraphics[width=1\linewidth]{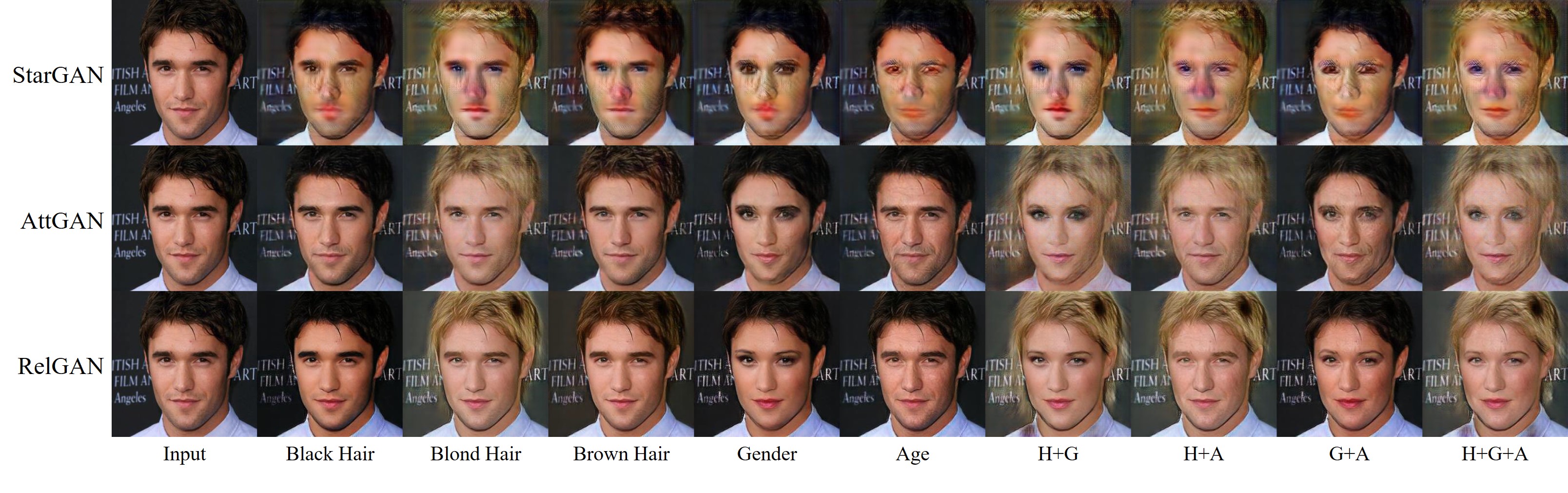}
    \caption{
    Facial attribute transfer results of StarGAN, AttGAN, and \ourGAN.
    Please zoom in for more details.
    }
    \label{fig:model_compare_2}
\end{figure}

\begin{figure}
    \centering
    \includegraphics[width=0.9\linewidth]{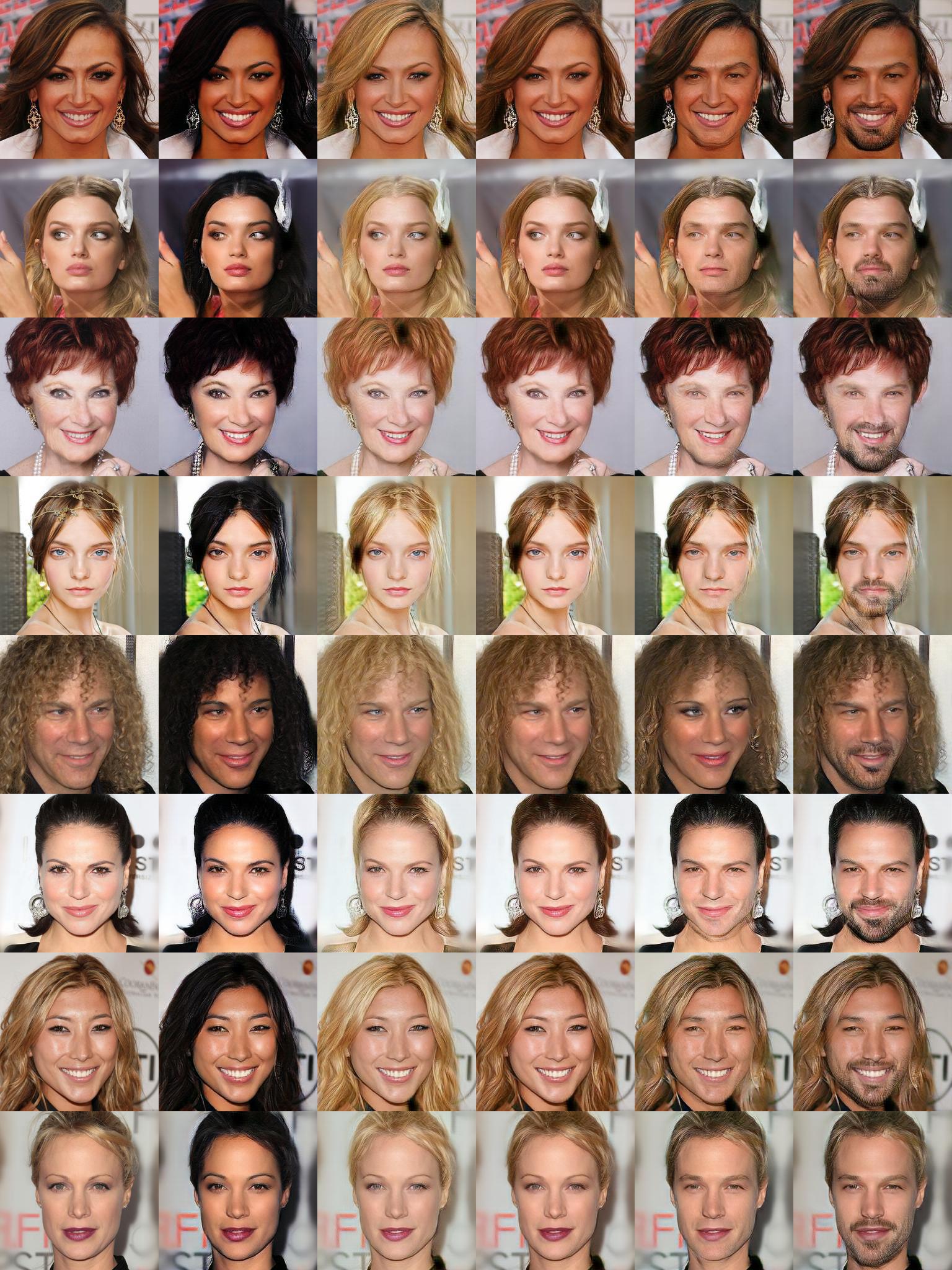}
    \caption{Single attribute transfer results.
    From left to right: input, black hair, blond hair, brown hair, gender, and mustache.
    }
    \label{fig:add_1}
\end{figure}

\begin{figure}
    \centering
    \includegraphics[width=0.9\linewidth]{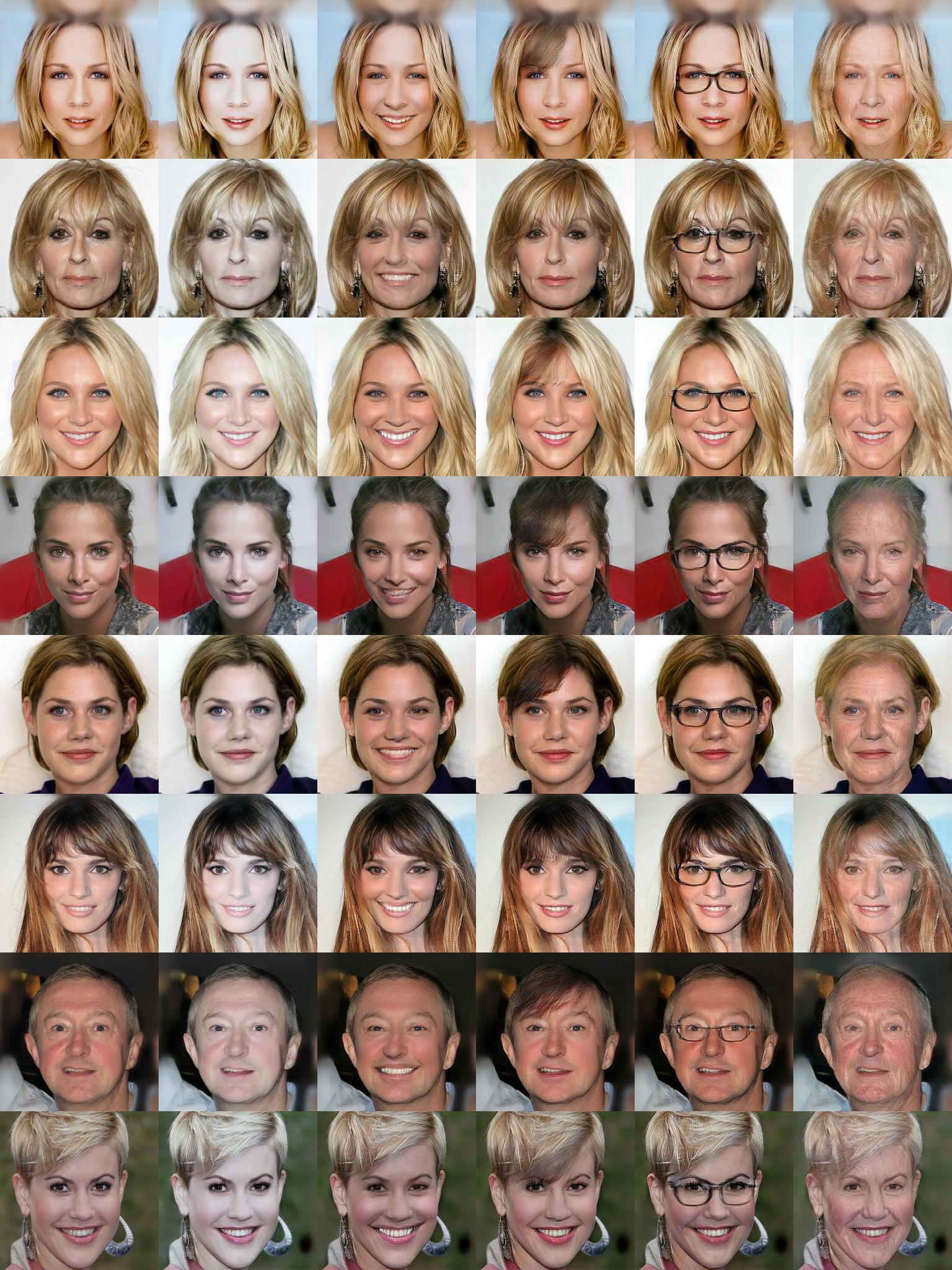}
    \caption{Single attribute transfer results.
    From left to right: input, pale skin, smiling, bangs, glasses, and age.
    }
    \label{fig:add_1p}
\end{figure}

\begin{figure}
    \centering
    \includegraphics[width=0.85\linewidth]{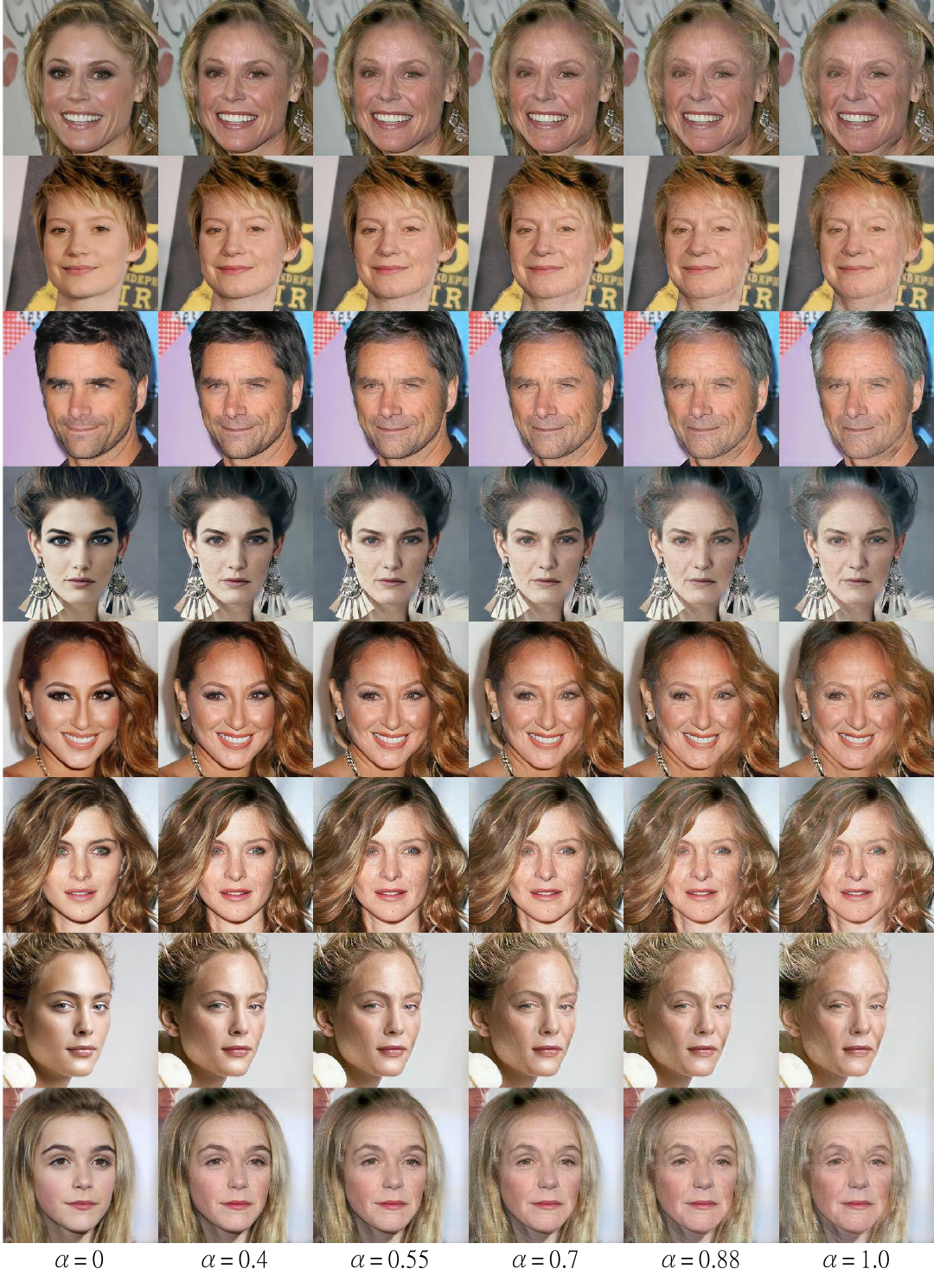}
    \caption{Single attribute interpolation results (age).
    \ourGAN~generates different levels of attribute transfer by varying $\alpha$.}
    \label{fig:inter_1}
\end{figure}

\begin{figure}
    \centering
    \includegraphics[width=0.85\linewidth]{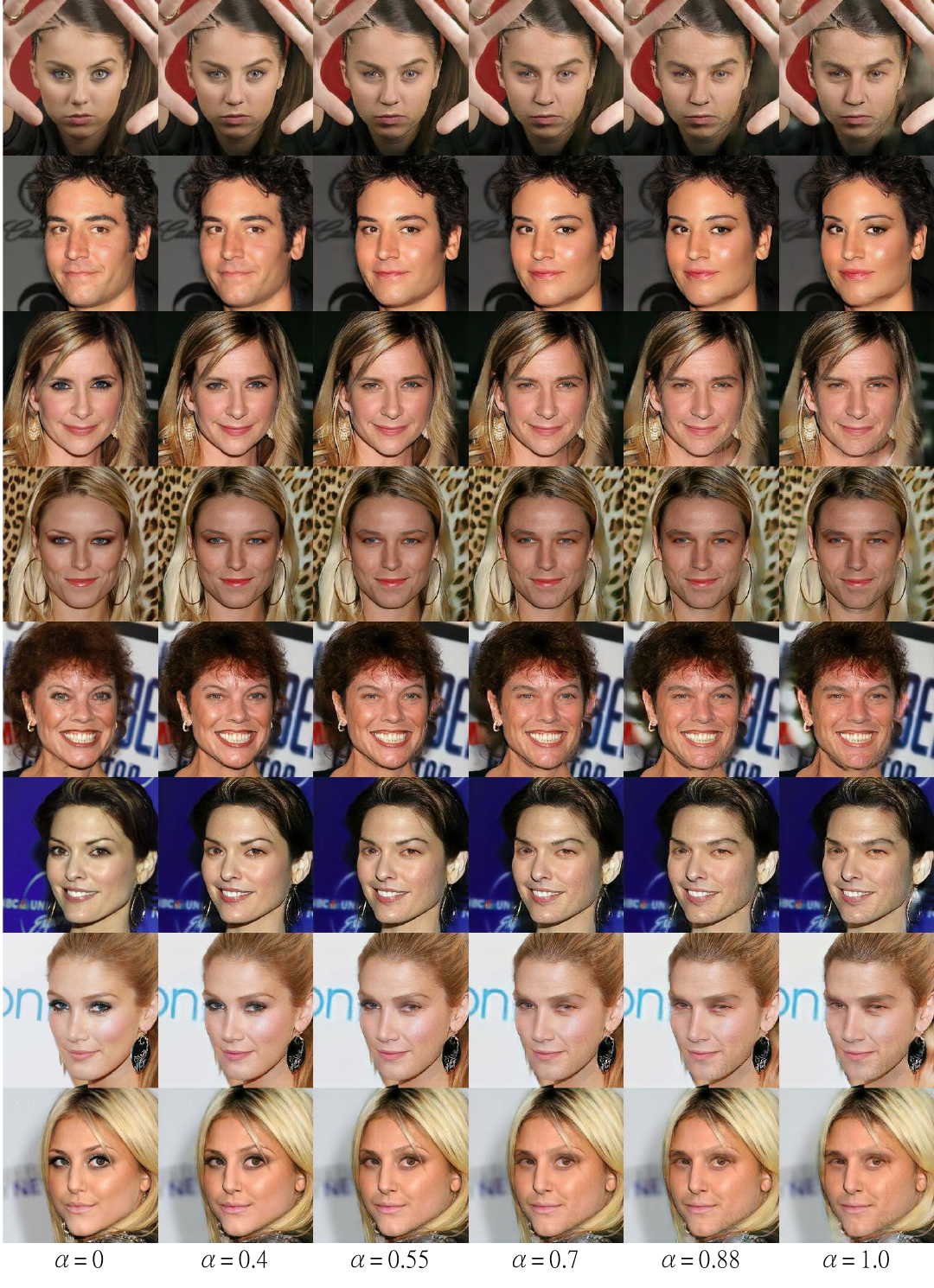}
    \caption{Single attribute interpolation results (gender).
    \ourGAN~generates different levels of attribute transfer by varying $\alpha$.}
    \label{fig:inter_2}
\end{figure}

\begin{figure}
    \centering
    \includegraphics[width=0.85\linewidth]{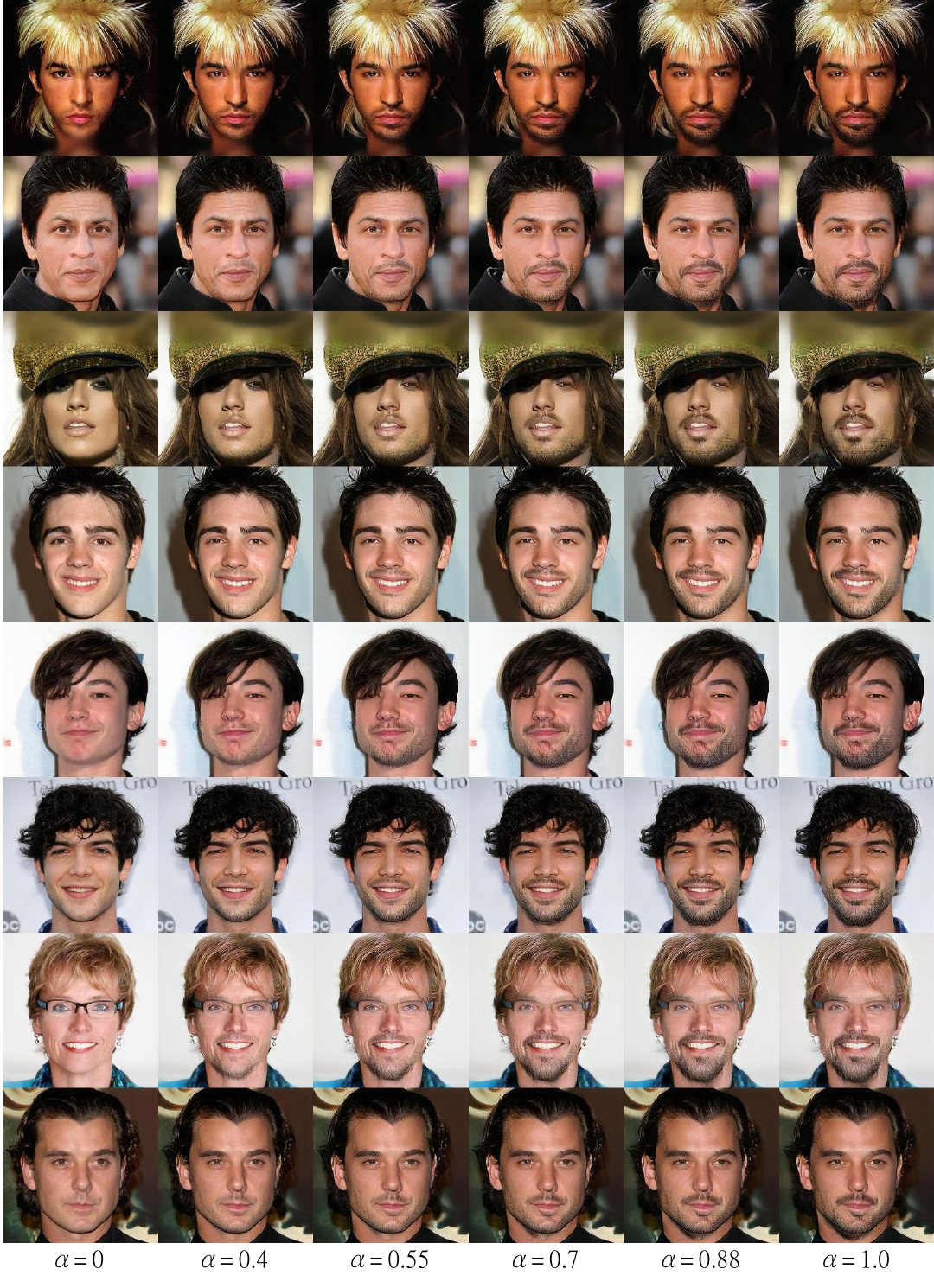}
    \caption{Single attribute interpolation results (mustache).
    \ourGAN~generates different levels of attribute transfer by varying $\alpha$.}
    \label{fig:inter_3}
\end{figure}

\begin{figure}
    \centering
    \includegraphics[width=0.85\linewidth]{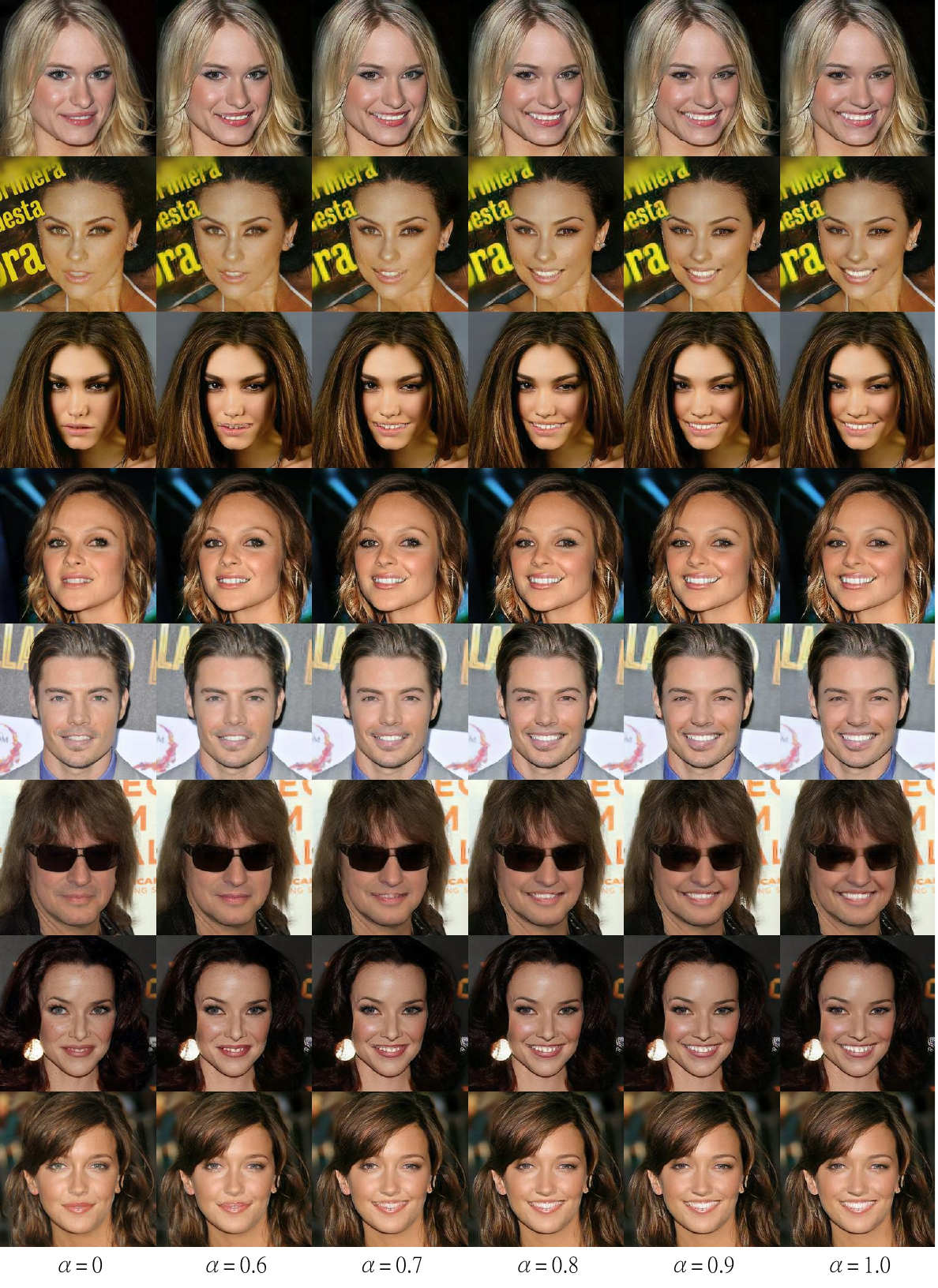}
    \caption{Single attribute interpolation results (smile).
    \ourGAN~generates different levels of attribute transfer by varying $\alpha$.}
    \label{fig:inter_4}
\end{figure}

Figure~\ref{fig:residual1} and~\ref{fig:residual2} show comparison results between StarGAN, AttGAN, and \ourGAN~on the hair color tasks.
The residual heat maps show that \ourGAN~preserves the smile attribute while the other methods strengthen the smile attribute.
Figure~\ref{fig:model_compare} and~\ref{fig:model_compare_2} show more comparison results.
Figure~\ref{fig:add_1},~\ref{fig:add_1p} show additional results on facial attribute transfer.
Figure~\ref{fig:inter_1},~\ref{fig:inter_2},~\ref{fig:inter_3}, and~\ref{fig:inter_4} show additional results on facial attribute interpolation. All the input images are from the CelebA-HQ dataset.

\vspace{-0.4cm}

\vspace{-0.4cm}

\vspace{-0.4cm}

\end{appendices}
\end{document}